\documentclass[lettersize,journal]{IEEEtran}
\usepackage{amsmath,amsfonts}
\usepackage{algorithmic}
\usepackage{algorithm}
\usepackage{array}
\usepackage[caption=false,font=normalsize,labelfont=sf,textfont=sf]{subfig}
\usepackage{textcomp}
\usepackage{stfloats}
\usepackage{url}
\usepackage{verbatim}
\usepackage{graphicx}
\usepackage{cite}
\usepackage{booktabs}
\usepackage{float}
\usepackage{makecell}
\usepackage{multirow}
\usepackage{multicol}
\usepackage{color}
\usepackage[colorlinks,
            linkcolor=blue,
            anchorcolor=blue,
            citecolor=blue]{hyperref}
\usepackage{bbm}
\usepackage{orcidlink}
\usepackage{float}
\usepackage{placeins} 
\usepackage{threeparttablex}

\begin{document}

\title{CLIP-Guided Adaptable Self-Supervised Learning for \\ Human-Centric Visual Tasks}


\author{Mingshuang Luo, Ruibing Hou\textsuperscript{$\dagger$ }, Bo Chao, Hong Chang,~\IEEEmembership{Member, IEEE}, \\ Zimo Liu\textsuperscript{$\dagger$ }, Yaowei Wang, Shiguang Shan,~\IEEEmembership{Fellow, IEEE}
\thanks{$\dagger$ Corresponding author.}
\thanks{Mingshuang Luo, Bo Chao, Hong Chang, and Shiguang Shan are with Key Laboratory of Intelligent Information Processing, Institute of Computing Technology (ICT), Chinese Academy of Sciences (CAS), and University of Chinese Academy of Sciences. (e-mail: \{mingshuang.luo, bo.chao\}@vipl.ict.ac.cn, \{changhong, sgshan\}@ict.ac.cn). Ruibing Hou is with Key Laboratory of Intelligent Information Processing, Institute of Computing Technology (ICT), Chinese Academy of Sciences (CAS). (e-mail: houruibing@ict.ac.cn).}
\thanks{Zimo Liu and Yaowei Wang are with Peng Cheng Laboratory. (e-mail: liuzm@pcl.ac.cn, wangyaowei@hit.edu.cn)}}

\markboth{Journal of \LaTeX\ Class Files,~Vol.~14, No.~8, August~2021}%
{Shell \MakeLowercase{\textit{et al.}}: A Sample Article Using IEEEtran.cls for IEEE Journals}


\maketitle

\begin{abstract}
Human-centric visual analysis plays a pivotal role in diverse applications, including surveillance, healthcare, and human-computer interaction. With the emergence of large-scale unlabeled human image datasets, there is an increasing need for a general unsupervised pre-training model capable of supporting diverse human-centric downstream tasks.  
To achieve this goal, we propose CLASP (CLIP-guided Adaptable Self-suPervised learning), a novel framework designed for  unsupervised pre-training in human-centric visual tasks.
CLASP leverages the powerful vision-language model CLIP to generate both low-level (\textsl{e.g.} body parts) and high-level (\textsl{e.g.} attributes) semantic pseudo-labels. These multi-level semantic cues are then integrated into the learned visual representations, enriching their expressiveness and generalizability. Recognizing that different downstream tasks demand varying levels of semantic granularity, CLASP incorporates a Prompt-Controlled Mixture-of-Experts (MoE) module. MoE 
dynamically adapts feature extraction based on task-specific prompts, mitigating potential feature conflicts and enhancing transferability. Furthermore, CLASP employs a multi-task pre-training strategy, where part- and attribute-level pseudo-labels derived from CLIP guide the representation learning process.
 Extensive experiments across multiple benchmarks demonstrate  that CLASP consistently outperforms existing unsupervised pre-training methods, advancing the field of human-centric visual analysis.
\end{abstract}


\section{Introduction}
\label{sec:intro}

\IEEEPARstart{H}{uman}-centric visual analysis plays a crucial role in various applications, including video surveillance \cite{ge2020mutual, ge2020self, zheng2015scalable, hou2020iaunet}, autonomous driving \cite{dollar2011pedestrian, mao2017can, chu2020detection}, and augmented reality \cite{fang2017rmpe, zhang2019fast}. This field encompasses a wide range of tasks, such as person re-identification (reID) \cite{gong2011person, luo2019bag, wang2020faster}, human parsing \cite{li2020self, wang2020hierarchical, ji2020learning}, pose estimation \cite{zheng2023deep, song2021human}, pedestrian detection \cite{liu2019high, gawande2020pedestrian}, and attribute recognition \cite{lin2019improving, wang2022pedestrian}. Despite significant progress in each individual task, most existing models are tailored for single task, leading to substantial overhead in model design and pre-training. To facilitate efficient and scalable deployment in real-world scenarios, it is essential to develop a unified human-centric pre-training model capable of effectively adapting to a wide range of downstream tasks.  

Human-centric pre-training typically follows two mainstream paradigms: supervised multi-task pre-training and unsupervised pre-training. 
The \textbf{supervised multi-task pre-training} paradigm leverages large-scale \textbf{annotated} datasets to jointly learn transferable representations across multiple tasks. For example, HCMoCo \cite{hong2022versatile} simultaneously optimizes for both pose estimation and human parsing, encouraging the model to capture shared structural semantics.  UniHCP \cite{ci2023unihcp} further enhances inter-task parameter sharing by employing a shared decoder equipped with task-specific queries. Extending to multimodal settings, Hulk \cite{wang2023hulk} proposes a generalist model that integrates 2D vision, 3D vision, skeleton-based representations, and vision-language tasks within a single framework.
In contrast, \textbf{unsupervised pre-training} aims to learn generalizable  representations from large-scale \textbf{unlabeled} data, providing a scalable and cost-effective alternative to supervised approaches.  Earlier studies \cite{luo2021self, yang2022unleashing} directly adopt DINO \cite{caron2021emerging} for self-supervised representation learning in human-centric scenarios. SOLIDER \cite{chen2023beyond} further introduces a semantic controller to regulate appearance and semantic information across tasks, while HAP \cite{yuan2023hap}  incorporates human part priors into the masked autoencoding (MAE) frameworks \cite{he2022masked}.
Given the superior scalability and reduced reliance on manual annotations, our goal is to develop an unsupervised human-centric pre-training model that fully leverages unlabeled data to support a wide range of downstream tasks.

\begin{figure}
	\centering
	\setlength{\abovecaptionskip}{0.1cm}
	\includegraphics[width=0.47\textwidth]{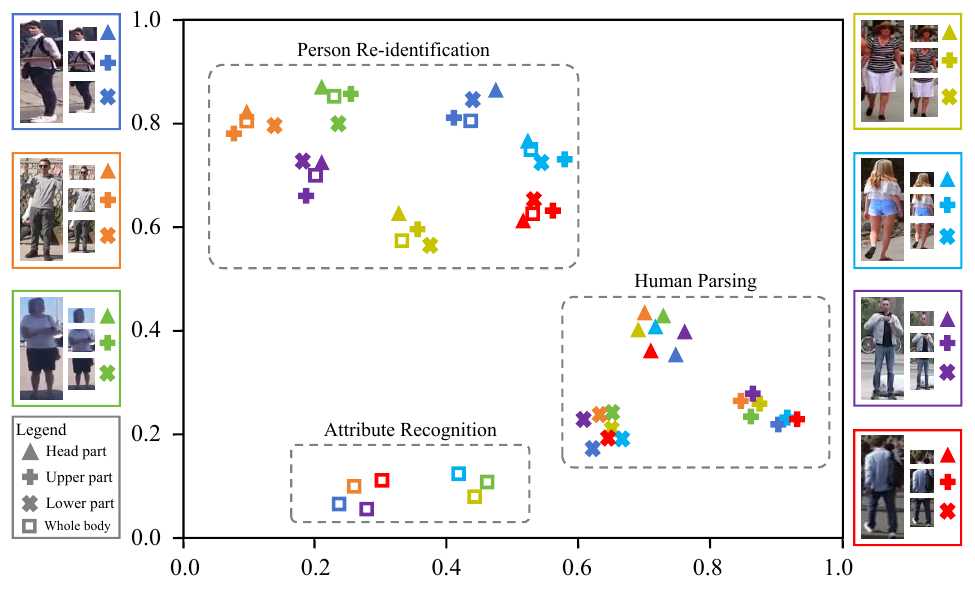}
	\caption{Feature representation space for person re-identification, attribute recognition, and human parsing tasks. Each task is trained independently using Swin-Tiny as the backbone, with features extracted via average pooling. The visualization highlights that different tasks demand distinct feature representations, highlighting the challenge of learning a unified representation across diverse human-centric tasks.}
    \vspace{-0.3cm}
	\label{different_tasks_representation_space} 
\end{figure}

Despite recent advances, developing a general-purpose unsupervised pre-training model for human-centric tasks remains highly challenging, primarily due to the heterogeneous representation requirements inherent to different tasks. As illustrated in Fig.~\ref{different_tasks_representation_space}, each task demands distinct feature characteristics. For instance, person reID relies heavily on \textbf{global appearance consistency}, requiring the preservation of discriminative visual cues to distinguish between individual identities. In contrast, semantic understanding tasks, such as human parsing, pose estimation and attribute recognition, require \textbf{multi-level semantic consistency} to capture fine-grained human-centric details. More specifically, human parsing and pose estimation emphasize  \textit{low-level part-aware semantics}, necessitating precise localization and differentiation of body parts. Meanwhile, attribute recognition depends on \textit{high-level global semantics} to infer abstract human attributes, such as gender or age. 

However, as highlighted  in \cite{chen2023beyond, yuan2024hap},  mainstream unsupervised pre-training approaches, such as DINO \cite{caron2021emerging} and MAE \cite{he2022masked}, struggle to learn universally transferable representations across diverse human-centric tasks. Specifically, DINO \cite{caron2021emerging} primarily captures appearance-based representations, which are well-suited for person reID task but insufficient for semantic understanding tasks that require rich multi-level semantic cues.
On the other hand, MAE \cite{he2022masked}, with its high masking ratio and reconstruction-based objective, tends to emphasize low-level visual features, thereby limiting its  capacity to model high-level semantics and global appearance cues.
SOLIDER \cite{chen2023beyond} makes partial progress by introducing  coarse-grained pseudo labels (\textsl{e.g.}, upper body, lower body, and shoes) to provide semantic supervision during DINO pre-training. However, it still faces two fundamantal limitations: (1) It entangles appearance and coarse semantic representations, lacking the ability to model fine-grained semantics (e.g., hat, hair) and high-level concepts (e.g., gender, age);  (2) It relies on manually tuned external parameters for adapting to each downstream task, which not only increases the adaptation cost but also limits scalability and flexibility in real-world deployments.

In this work, we propose CLASP (\textbf{CL}IP-guided \textbf{A}daptable \textbf{S}elf-su\textbf{P}ervised Learning), a unified unsupervised pre-training framework designed to address the  heterogeneous representation requirements across diverse human-centirc  tasks.
\textbf{First,} CLASP leverages the powerful vision-language model CLIP to automatically 
 generate \textit{multi-level semantic pseudo labels}, covering both \textit{low-level body parts} and \textit{high-level human attributes}. These pseudo labels supervise two pretext tasks—body part classification and attribute classification—guiding the model to acquire rich and robust semantic representations across different levels.
\textbf{Second},  to alleviate potential feature conflicts among different tasks, we introduce a Prompt-Controlled Mixture-of-Experts (\textbf{PC-MoE}) module. This module comprises a set of expert sub-networks, a channel-wise gating mechanism, and a global-wise gating mechanism, all dynamically modulated by a task-specific learnable prompt.  The prompt serves as a control signal to  steer the expert selection process, enabling adaptive and task-aware representation learning. 
\textbf{Finally}, CLASP is optimized through a set of hierarchical supervision signals that jointly supervise both appearance features and multi-level semantic cues.  
This combination of prompt-guided expert control and multi-level joint supervision empowers the pre-trained model with strong generalization and adaptability across a wide range of downstream human-centric tasks.

Through comprehensive  quantitative analysis, we demonstrate  that CLIP can effectively generate reliable semantic pseudo labels for both human attributes and body parts. In addition, qualitative results show that task-specific prompt embeddings in CLASP successfully guide the model to learn adaptable representations. Extensive experiments on six diverse human-centric downstream tasks validate the effectiveness of our framework: CLASP achieves significant improvements on semantic tasks, while maintaining competitive performance on appearance-centric tasks. These results highlight the versatility and generalization capability of CLASP across heterogeneous human-centric scenarios.
\section{Related Work}
\label{sec:formatting}

\subsection{Human-centric Visual Perceptions}
Human-centric visual tasks, such as person reID \cite{ge2020mutual, ge2020self, ye2021deep, hou2020iaunet}, human parsing \cite{ zhang2022human, e2ehumanparsing}, attribute recognition \cite{wang2022pedestrian, visiontransformerAR}, pose estimation \cite{zhang2019fast, zheng2023deep}, person search \cite{li2017person, lan2018person}, and pedestrian detection \cite{mao2017can, chu2020detection}, have witnessed remarkable progress in recent years. However, most existing methods tackle each task in isolation, relying on specifically annotated datasets designed for individual objectives. 
Recently, there has been a growing shift towards unified human-centric visual perception modeling, which aims to exploit shared representations across multiple tasks. Some studies  \cite{ci2023unihcp, tang2023humanbench, hong2022versatile} explore fully supervised pre-training for multi-task learning, while others \cite{chen2023beyond} utilize large-scale unlabeled pedestrian datasets to learn general-purpose visual backbones via self-supervised learning. Additionally, several works extend this line of research to 3D domain, focusing on 3D human-centric pre-training \cite{chen2022liftedcl} and unified motion modeling \cite{zhu2022motionbert}.

In this paper, we follow a self-supervised learning paradigm, leveraging large-scale unlabeled pedestrian data to learn robust and transferable feature representations. Our goal is to develop a versatile visual front-end capable of generalizing across a wide range of downstream human-centric visual tasks.

\vspace{-0.3cm}
\subsection{Self-supervised Representation Learning}
Self-supervised learning has gained prominence in visual representation learning, with prominent frameworks including contrastive learning \cite{caron2021emerging, chen2020improved}, masked image modeling (MIM) \cite{he2022masked}, and clustering-based methods \cite{el2021large, zhou2021ibot}. These approaches can be broadly categorized into two groups: (1) Contrastive learning methods, such as SimCLR \cite{chen2020simple} and DINO \cite{caron2021emerging}, which learn discriminative features by distinguishing positive and negative sample pairs. (2) Masked image modeling techniques, such as MAE \cite{he2022masked}, which aims to reconstruct masked image patches to promote holistic feature learning.

Recent studies have applied self-supervised learning to human-centric tasks.  For instance, HumanBench \cite{tang2023humanbench} explore both contrastive learning and MAE for human-centric perception, and observe that MAE performs suboptimally in reID tasks. In contrast, SOLIDER \cite{chen2023beyond} demonstrates that DINO-based pre-training achieves competitive performance across a variety of downstream human-centric tasks. Motivated by its strong generalization capabilities, we adopt DINO as the backbone for our self-supervised learning framework.

\begin{figure*}[htp]
	\centering
	\setlength{\abovecaptionskip}{0.1cm}
	\includegraphics[width=1.00\textwidth]{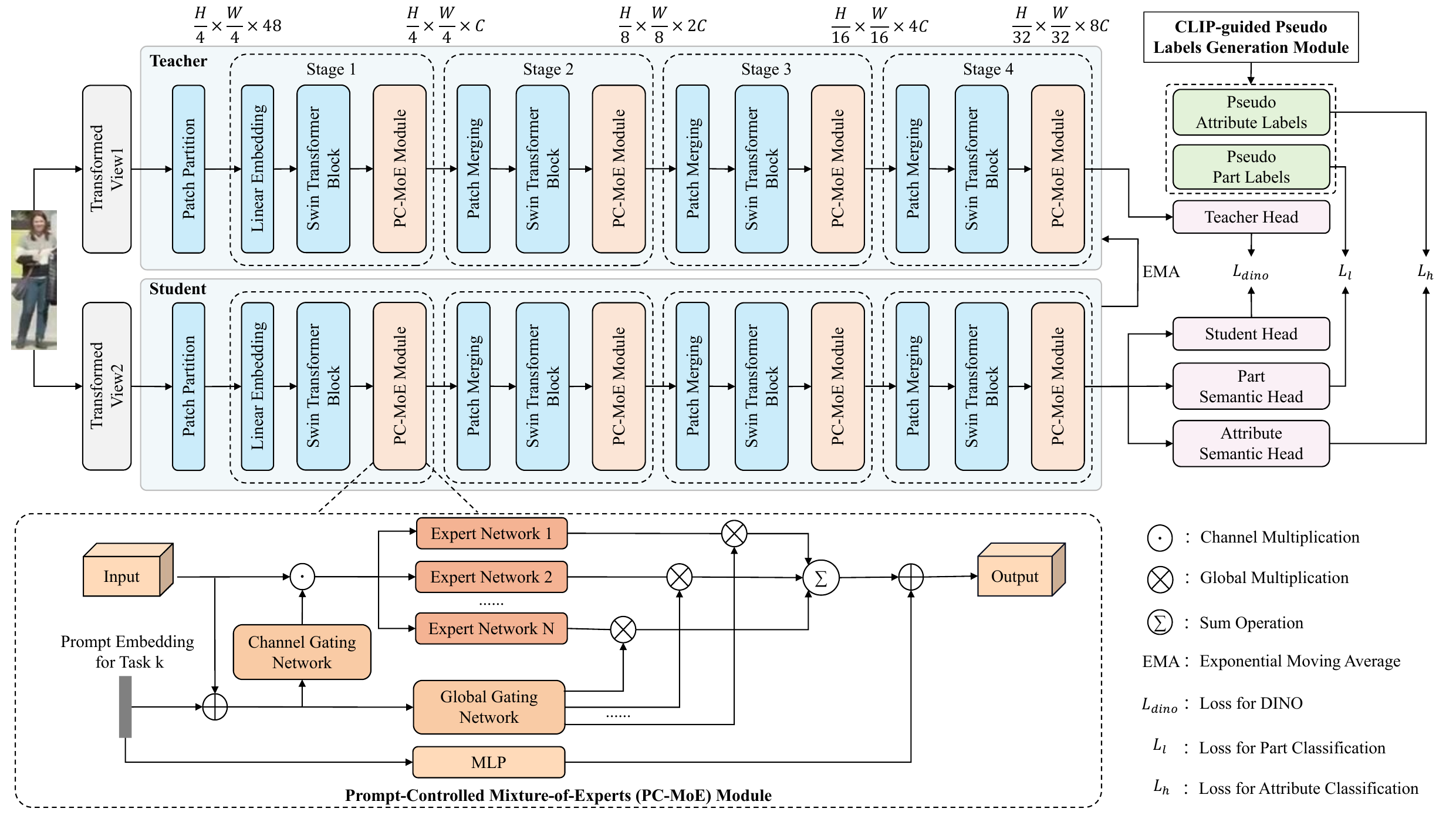}
	\caption{The overall architecture of CLASP framework. CLASP consists of four key components: (1) a teacher branch, which generates feature representations using a Prompt-Controlled Mixture of Experts (PC-MoE) module, updated via EMA; (2) a student branch with an identical structure, which learns by distilling knowledge from the teacher; (3) a CLIP-guided Pseudo Labels Generation Module, responsible for producing pseudo part-level and attribute-level semantic labels; (4) three task-specific heads, each designed to optimize a distinct pre-training objective.}
	\label{overview_model_fig} 
    \vspace{-0.2cm}
\end{figure*}

\subsection{Language-driven Visual Perception Modeling}
Language-driven vision models (VLMs) have demonstrated remarkable generalization across various vision tasks, including zero-shot classification \cite{esmaeilpour2022zero, novack2023chils}, open-vocabulary object detection \cite{wu2023aligning, shi2023edadet}, and semantic segmentation \cite{he2023clip}. Foundational works such as CLIP \cite{radford2021learning}, PaLI-3 \cite{chen2023pali}, and RegionCLIP \cite{zhong2022regionclip} have demonstrated the effectiveness of leveraging large-scale image-text pairs to learn joint visual-semantic representations.

Building on this, recent efforts have explored VLMs in human-centric tasks. For instance, CLIP-S$^4$ \cite{he2023clip} integrates CLIP-based semantic guidance into self-supervised learning to enforce semantic consistency. In this work, we harness VLMs for weakly supervised label generation, extracting part-level and attribute-level semantic cues from pedestrian images. These cues serve as soft supervision signals to guide our self-supervised feature learning process.
\vspace{-0.3cm}
\subsection{Mixture of Experts}
The Mixture of Experts (MoE) framework \cite{jacobs1991adaptive} enables dynamic model specialization through conditional computation. A typical MoE consists of a gating network that selectively routes inputs to a subset of expert networks,  allowing for efficient use of model capacity. MoE has been successfully applied across a range of domains, including natural language processing \cite{yi2023edgemoe, dou2024loramoe}, computer vision \cite{chen2023adamv, lin2024moe}, and speech recognition \cite{hu2023mixture, song2024u2++}. It has also shown promise in multimodal and multi-task learning \cite{li2024cumo, li2024uni}. Recent work has focused on improving the efficiency and scalability of MoE architectures. For instance, DeepSeekMoE \cite{dai2024deepseekmoe} introduces fine-grained expert partitioning to enhance specialization with reduced computational overhead. Similarly, Vision MoE (V-MoE) \cite{riquelme2021scaling} adapts MoE paradigm to Vision Transformers, achieving state-of-the-art image recognition performance with sparse computation.
\section{Method}
The overall architecture of the proposed CLASP is illustrated in Fig.~\ref{overview_model_fig}. 
In section \ref{sec:3.1}. we present the process of generating multi-level semantic pseudo labels using the vision-language model CLIP \cite{radford2021learning}.
In section \ref{sec:3.2}, we introduce the Prompt-Controlled Mixture-of-Experts (PC-MoE) module, which is designed to alleviate feature conflicts across tasks with varying levels of semantic granularity, thereby enabling more adaptive and task-aware representation learning. In section \ref{sec:3.3}, we provide the overall loss function to train the pre-training model.

\vspace{-0.3cm}
\subsection{CLIP-guided Multi-level Semantic Supervision}
\label{sec:3.1}
 DINO \cite{caron2021emerging} has been widely adopted for visual representation learning and has shown promising performance on various human-centric tasks \cite{chen2023beyond,yang2022unleashing}. As a contrastive learning framework, DINO primarily captures global appearance features, which makes it effective for appearance-oriented tasks such as person reID. However, it lacks the capacity to model fine-grained and high-level semantic information, thereby limiting its effectiveness in semantic  tasks, such as human parsing, pose estimation, and attribute recognition. 
 In this work, we adopt DINO \cite{caron2021emerging} as our baseline framework and seek to enrich its representation with multi-level semantic cues, enhancing its suitability for semantic human-centric downstream tasks.

Importantly, different human-centric tasks require semantic understanding at different levels of granularity.   For instance, human parsing \cite{li2020self} and pose estimation \cite{song2021human} rely on \textit{low-level} semantic cues, such as head, arms, and shoes, to accurately localize and segment body parts. In contrast, attribute recognition \cite{lin2019improving} depends on high-level semantic concepts, such as gender and age.  This diversity highlights the necessity for self-supervised human-centric representation learning to inherently capture multi-level semantic information. Motivated by the strong semantic encoding capabilities of vision-language models, we leverage CLIP \cite{radford2021learning} to automatically generate both low-level part-based pseudo labels and high-level attribute pseudo labels, which serve as supervision signals to enhance semantic representation learning within our pre-training framework. 

\begin{figure*}
	\centering
	\setlength{\abovecaptionskip}{0.1cm}
	\includegraphics[width=1.00\textwidth]{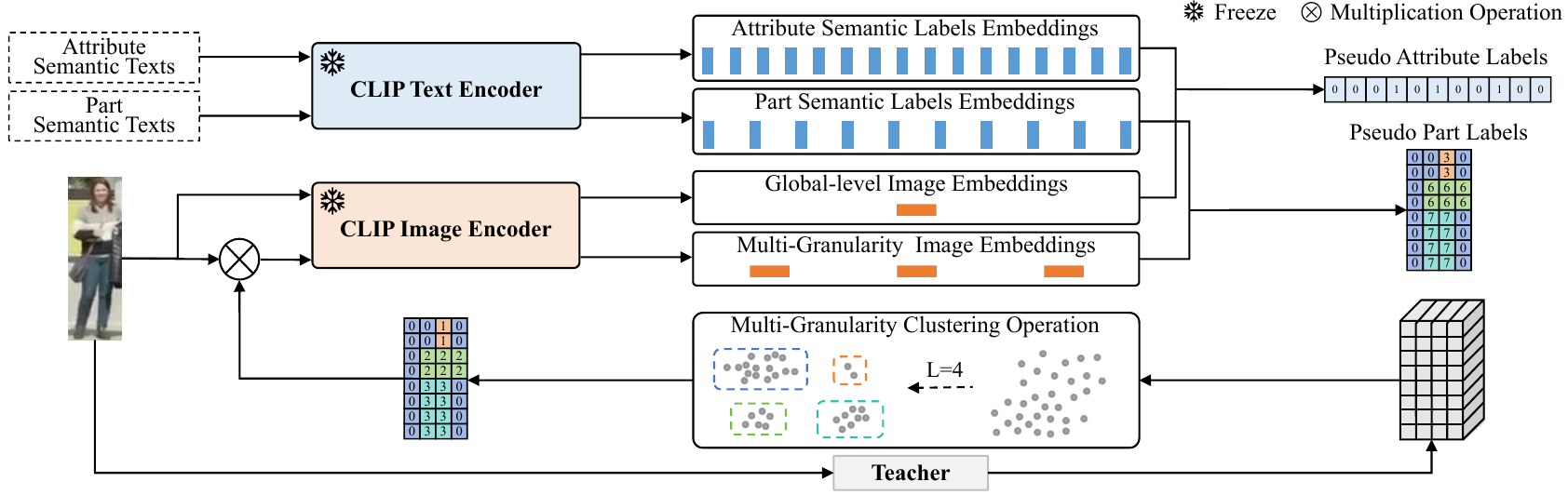}
	\caption{Illustration of the CLIP-guided Pseudo Label Generation Module. This module aims to generate pseudo semantic labels at multiple granularity, including part-level labels and global attribute-level labels, by leveraging CLIP's powerful image-text alignment capability. $L=4$ is an example of the variable sampled from a predefined set $S=\{2,3,4\}$.}
    \vspace{-0.3cm}
	\label{clip-pseudo-module} 
\end{figure*}

\

\noindent
\textbf{CLIP-Guided Multi-Granularity Part Semantic Labels.} \
As illustrated  in Fig.~\ref{clip-pseudo-module}, this branch focuses on generating \textbf{part-level pseudo semantic labels} guided by CLIP.  We begin by performing a clustering operation to segment image pixels into distinct body regions, which are then assigned semantic labels corresponding to specific body parts (\textsl{e.g.} head, arms, shoes) using CLIP-based similarity matching. Motivated by observation in \cite{chen2023beyond}, which shows that human DINO representations naturally exhibit part-level structrual separation, we leverage features extracted by a pre-trained DINO model to guide the clustering process. 

Specifically, given an unlabeled person image $\boldsymbol{I} \in \mathbb{R}^{3 \times H \times W}$, we first apply the DINO teacher encoder to extract its feature representation $\boldsymbol{F} \in \mathbb{R}^{c \times h \times w}$.  The resulting feature map $\boldsymbol{F}$ can be viewed as a set of $h\times w$ token vectors, each of dimension $c$. Following \cite{chen2023beyond}, we initially perform K-means clustering \cite{hartigan1979algorithm} on these token vectors to separate them into two coarse categories: foreground and background. Here, we also introduce a dual-criteria filtering mechanism to ensure both spatial and semantic integrity of the human image: (i) \textbf{Spatial Filtering}: These tokens (from DINO features) are clustered into foreground (\emph {human}) and background categories. Only images with over $50\%$ human tokens are retained to ensure human-centric focus.  (ii) \textbf{Semantic Filtering:}  The retained images undergo a zero-shot quality check via CLIP using the prompt ``a photo of a person”. Only images with a cosine similarity above $0.9$ to the prompt embedding are preserved.  This dual-filtering process guarantees that only high-quality, human-focused images are used for pseudo-label generation. 
To account for variations in visible body parts (foreground) due to occlusions and diverse camera viewpoints, we introduce a \textbf{Multi-Granularity Clustering Operation} to further segment the foreground region into semantically meaningful sub-parts. Specifically, we randomly sample a cluster number $L$ from a predefined candidate set $\mathbb{S}$, and apply K-means clustering with $L$ clusters on the foreground tokens. This process yields $L$ spatial region masks $\left\{\boldsymbol{M_i} \in \mathbb{R}^{h \times w}\right\}_{i=1}^L$, where each $\boldsymbol{M_i}$ is a binary mask indicating the spatial location of the $i$-th part-level region within the foreground area. 

Next, leveraging the region masks $\left\{\boldsymbol{M_i}\right\}_{i=1}^L$, we utilize CLIP to assign part-level pseudo-label to each segmented region. Specifically, each region mask $\boldsymbol{M_i}$ is first resized to the original image resolution $H\times W$. The resized mask is then applied to the input image $I$ to isolate the corresponding semantic region:
\begin{equation} 
\boldsymbol{\widetilde{I}_i} = \boldsymbol{I} \odot \mathcal{R}\left(\boldsymbol{M_i}\right), 
\end{equation} 
where $\mathcal{R}\left(\cdot\right)$ denotes the resizing operation and $\odot$ represents element-wise multiplication. The resulting masked image $\boldsymbol{\widetilde{I}_i}$ retains only the visual content associated with the $i$-th region. The masked region is then fed into CLIP image encoder to obtain its visual representation  $\boldsymbol{z_i} \in \mathbb{R}^d$. To assign a semantic label, we compute the cosine similarity between $\boldsymbol{z_i}$ and a set of predefined textural descriptions $\left\{\boldsymbol{t_j} \in \mathbb{R}^d \right\}_{j=1}^K$, where each $\boldsymbol{t_j}$ is the CLIP-encoded text embedding corresponding to a candidate body part (\textsl{e.g.}, \textit{head, upper body, shoes}). The part-level pseudo-label for region $i$ is determined by selecting the text with the highest similarity:
\begin{equation}
q^{p}_i = \mathop{\arg\max}_{j \in \left\{1,2,\dots, K\right\}} \ \cos \left(\boldsymbol{z_i}, \boldsymbol{t_j}\right),
\end{equation} 
where $\cos$ denotes cosine similarity.

Finally, the predicted part-level pseudo labels $\left\{q^{p}_i\right\}_{i=1}^L$ are propagated to the original image pixels, resulting in pixel-level part label map $\boldsymbol{Q^p} \in \mathbb{R}^{H\times W}$. Specifically, each pixel at location $\left(h,w\right)$ is assigned as follows:
\begin{equation}
\left[\boldsymbol{Q^p}\right]_{\left(h,w\right)} = \left\{
\begin{aligned}
&q^p_i, \quad \text{if} \ \ \mathcal{R}\left(M_i\right)_{h,w}=1 \\
&0, \quad \text{if} \ \ \prod \limits_{j=1}^n \mathcal{R}\left(M_j\right)_{h,w} = 0.
\end{aligned}	
\right.
\end{equation}
Here, label $0$ denotes background pixels that are not covered by any region mask. The resulting pixel-level label map $\boldsymbol{Q^p}$ provides dense semantic supervision, and is used to train a part classification head as one of the pretext tasks in our self-supervised learning framework.

\

\noindent
\textbf{CLIP-Guided Attribute Semantic Labels.} \
In this part, we aim to generate pseudo-labels for high-level human attributes by leveraging the semantic alignment capabilities of CLIP. To this end, we first define a set of common human attributes $\left\{a^i\right\}_{i=1}^{M}$, such as \textit{gender}, \textit{hairstyle} and \textit{age}, and seek to assign  pseudo-label to each attribute $a^i$ based on visual-textual similarity. We also apply the dual-criteria filtering mechanism to human images to reduce the interference of the image noise.
Formally, as shown in Fig.~\ref{clip-pseudo-module}, given an unlabeled person image $\boldsymbol{I}$, we extract its global visual representation $\boldsymbol{z}$ using the CLIP image encoder. For each attribute $a^i$, we construct a  prompt template to describe its candidate values in natural language. For example, for the \textit{gender} attribute, we adopt the template: `$\mathrm{A}$ $\mathrm{photo}$ $\mathrm{of}$ $\mathrm{a}$ $\mathrm{\left\{attribute \ label\right\}}$ $\mathrm{person.}$''. Similarly, for the \textit{hairstyle} attribute, we use the template: $\mathrm{A}$ $\mathrm{photo}$ $\mathrm{of}$ $\mathrm{a}$ $\mathrm{person}$ $\mathrm{with}$ $\mathrm{\left\{ attribute \ label\right\}}$. 
Using these templates, we generate a set of textual descriptions for each attribute $a^i$. These descriptions are then encoded by the CLIP text encoder to obtain text feature vectors $\left\{\boldsymbol{w_k^i}\right\}_{k=1}^{K^i}$, where $K^i$ is the number of candidate labels for attribute $a^i$. 
To assign a pseudo-label, we compute the cosine similarity between the visual feature $\boldsymbol{z}$ and each text embedding $\boldsymbol{w_k^i}$, retain only those with similarity scores above a 0.5 threshold, and assign the label with the highest similarity:
\begin{equation}
q^a_i = \mathop{\arg\max}_{k\in \left\{1,2,\dots, K^i\right\}} \ \cos \left(\boldsymbol{z}, \boldsymbol{w^i_k}\right).
\end{equation}

Following common practice \cite{ci2023unihcp},  the attribute classification task is typically formulated as a set of independent binary classification task, one for each attribute. Accordingly, we convert each predicted attribute pseudo-label $q^a_i$ into a one-hot vector, denoted as $\boldsymbol{Q^a_i}=\mathrm{onehot}\left(q^a_i\right)$, where  $\mathrm{onehot}$ is the standard one-hot encoding function that maps a discrete attribute value to a binary vector. The one-hot pseudo-label $\left\{\boldsymbol{Q^a_i}\right\}_{i=1}^M$ serves as the supervisory signal for the corresponding attribute classification pretext task during training.

\vspace{-0.3cm}

\subsection{Prompt-Controlled MoE Module}
\label{sec:3.2}
As discussed above, an effective human-centric pre-training representation should encapsulate semantic information across multiple levels to support diverse downstream tasks. However, each task exhibits distinct representational demands\footnote{Notably, overall appearance information can be implicitly synthesized by integrating multi-level semantic features from human body parts and attributes.}. 
Nevertheless, existing approaches \cite{chen2023beyond} often entangle semantic features across different levels, making it challenging   for downstream tasks to  extract task-specific information. To address this, we turn to the  Mixture of Experts (MoE) architecture \cite{jacobs1991adaptive}, which inherently supports multi-perspective modeling and  adaptive feature learning. 

However, conventional MoE architecture has two key limitations: (1) The gating network relies solely on the model’s internal features, limiting external controllability and adaptability to diverse downstream tasks. (2) MoE designs typically apply coarse-grained, expert-level gating, overlooking the fine-grained and channel level control, a critical aspect for capturing nuanced semantic variations. 
To overcome these limitations, we propose a Prompt-Controlled Mixture of Experts (PC-MoE), a novel module within our CLASP framework. PC-MoE enables explicit disentanglement and dynamic utilization of hierarchical semantic representation, ensuring more effective task-adaptive feature extraction. 

\

\noindent
\textbf{PC-MoE structure.} \ 
The structure of our proposed PC-MoE module is illustrated in Fig.~\ref{overview_model_fig}. Compared to conventional MoE, PC-MoE introduces three key enhancements: (1) \textbf{Prompt-guided feature adaptation}: Task-specific prompt vectors are incorporated to guide the model towards learning task-adaptive representations;
(2) \textbf{Hierarchical gating mechanism}:  A dual-gating system combining channel-wise and global gating networks enables comprehensive feature control at both fine-grained and coarse levels;
(3) \textbf{Residual prompt enhancement:} Learnable task-specific prompts are utilized as residual vectors during the forward pass to augment the model's representational capacity.

As shown in Fig.~\ref{overview_model_fig}, we integrate the PC-MoE module into each stage of the CLASP framework. For stage $i$, the module takes as input an image feature map $\boldsymbol{F^i} \in \mathbb{R}^{c^i\times h^i \times w^i}$. For notational simplicity, we omit the stage superscript $i$ in the following. First, we introduce task-specific learnable prompt embedding as $\boldsymbol{e_k} \in \mathbb{R}^{c}$, where $k \in \left\{0,1,2\right\}$ indexes the pretraining tasks.  And we combine $\boldsymbol{e_k}$ with the input feature $\boldsymbol{F}$ through element-wise addition: $\boldsymbol{F_{\mathrm{gate}}}=\boldsymbol{F}\oplus\boldsymbol{e_k}$, where $\boldsymbol{F_{\mathrm{gate}}} \in \mathbb{R}^{c \times h \times w}$ denotes the prompt-augmented features. Secondly, the augmented feature $\boldsymbol{F_{\mathrm{gate}}}$ is then processed by two parallel gating networks, a channel-wise gating $\mathcal{G}_c$ and a global-wise gating $\mathcal{G}_g$, to generate modulation weights:
\begin{align}
    \mathcal{G}_{c}(\boldsymbol{F_{\mathrm{gate}}}) &= \text{Softmax}(\boldsymbol{F_{\mathrm{gate}}} \cdot W_c)),
\end{align}
\vspace{-0.4cm}
\begin{align}
    \mathcal{G}_{g}(\boldsymbol{F_{\mathrm{gate}}}) &= \text{TopK}\Big(\text{Softmax}(\boldsymbol{F_{\mathrm{gate}}} \cdot W_g))  \nonumber \\
         &\quad + \mathcal{N}(0, 1) \cdot \text{Softplus}(\boldsymbol{F_{\mathrm{gate}}} \cdot W_{\mathrm{noise}}) \Big).
\end{align} 
Here, the computing process of $\mathcal{G}_{g}$ is followed with the original MoE \cite{jacobs1991adaptive}, $N$ refers to the number of expert networks, $W_c \in \mathbb{R}^{c \times c}$ refers to the weights of the channel-wise gating network $\mathcal{G}_{c}$, $W_g \in \mathbb{R}^{c \times N}$ refers to the weights of the global-wise gating network $\mathcal{G}_{g}$, and $W_{\mathrm{noise}}$ refers to the weight of noise. $\text{TopK}(\cdot,j)$ sets all elements in the vector to zero except the elements with the largest $K$ values. Softplus is the smooth approximation to ReLU: $\text{Softplus}(x) = log(1+\text{exp}(x))$.

After obtaining the modulation weights through both channel-wise and global-wise gating networks, we apply these adaptive weights to dynamically combine the expert network outputs. The final representation of the PC-MoE module is computed as
\begin{align}
\boldsymbol{Y_{\mathrm{MoE}}} &= \sum_{j=1}^{N} \mathcal{G}^j_{g}\left(\boldsymbol{F_{\mathrm{gate}}}\right) \otimes \mathcal{E}^j\left(\boldsymbol{F} \odot \left(\mathcal{G}_{c}\left(\boldsymbol{F_{\mathrm{gate}}}\right)\right)\right)
\end{align} 
where $\mathcal{E}^j$ refers to the j-th expert network, $\otimes$ refers to the  multiplication operation, and $\odot$ refers to the dot product operation. 

To further enhance task-specific feature discrimination, we augment the MoE output with learnable task prompt, formulated as:
\begin{equation}
\boldsymbol{Y_{\mathrm{output}}} = \boldsymbol{Y_{\mathrm{MoE}}}  + \mathrm{FC}\left(\boldsymbol{e_k}\right),
\end{equation}
where $\mathrm{FC}$ refers to a full connected layer.
In the CLASP framework, PC-MoE modules are integrated into all stages of both teacher and student branches. At each stage, we employ distinct learnable task-specific prompt embeddings to guide and specialize the feature learning process.

\subsection{Loss Functions}
\label{sec:3.3}
In this section, we detail the optimization objectives for training CLASP.  CLSAP leverages part-level and attribute-level annotations derived from CLIP to supervise three  learning tasks. Specifically, our framework combines four losses: contrastive learning based on DINO, part-level classification, attribute-level classification and MoE loss. 

\textbf{DINO contrastive loss.} Let $ \boldsymbol{F_t} \in \mathbb{R}^{c \times h\times w}$ and $ \boldsymbol{F_s} \in \mathbb{R}^{c \times h\times w}$ denote the feature maps generated by the teacher and student branches, respectively. To derive a compact global representation, we apply  global average pooling operation (GAP) to both feature maps. The resulting pooled features are then projected into an embedding space using the teacher projection head $t_{\theta_{t}}$ and the student projection head $s_{\theta_{s}}$. The DINO contrastive loss $\mathcal{L}_{\mathrm{dino}}$ is formally defined:
\begin{equation}      
\mathcal{L}_{\mathrm{dino}} = \mathrm{DINOLoss}\left(t_{\theta_{t}}\left(\text{GAP}\left(\boldsymbol{F_t}\right)\right), s_{\theta_{s}}\left(\text{GAP}\left(\boldsymbol{F_s}\right)\right)\right),
\end{equation}
where $\mathrm{DINOLoss}$ is the DINO self-supervised loss defined in \cite{caron2021emerging}, and $\mathrm{GAP}$ denotes the global average pooling operation.

\textbf{Part-level classification loss.} For the part-level classification pre-training task, a part semantic head $h_{\theta_{l}}$ is introduced to classify each token vector in $\boldsymbol{F_{s}}$ according to the part label $Q^p$. The low-level semantic loss $\mathcal{L}_{l}$ is computed as the mean cross-entropy classification losses across all tokens:
\begin{equation}
\mathcal{L}_{l}=\frac{1}{w \times h} \sum_{\substack{h, w}}  \ell_{ce} \left(\left[\boldsymbol{F_{s}}\right]_{\left(h,w\right)}, \left[\boldsymbol{Q^p}\right]_{\left(h,w\right)}; h_{\theta_{l}} \right),
\end{equation}
where $\ell_{ce}$ denotes the cross-entropy function. 

\textbf{Attribute-level classification loss.} For attribute-level classification pre-training task, an attribute semantic head $h_{\theta_{h}}$ is employed to classify $\boldsymbol{F_s}$ based on a set of attribute labels $\left\{\boldsymbol{Q^a_i}\right\}_{i=1}^{M}$. Following \cite{ci2023unihcp}, the high-level attribute loss $\mathcal{L}_{h}$ is computed as the mean binary cross-entropy classification loss across all attribute values:
\begin{equation}
\mathcal{L}_h= \frac{1}{\sum_{i=1}^{M} K^i} \sum_{i=1}^{M} \sum_{j=1}^{K^i} \ell_{bce} \left(\mathrm{GAP}\left(\boldsymbol{F_s}\right), \left[\boldsymbol{Q^a_i}\right]_{j}; h_{\theta_{h}}\right),
\end{equation}
where $\ell_{bce}$ denotes the binary cross-entropy function, $K^i$ denotes the number of candidate labels for attribute $a^i$, and $\mathrm{GAP}$ denotes the global average pooling operation. 

\textbf{MoE loss.} The PC-MoE module employs a global gating network to route inputs to individual expert networks. Imbalanced routing can lead to expert overload or underused, compromising performance and resource efficiency. To mitigate this, we adopt the load balancing loss from \cite{shazeer2017outrageously} in each stage $i$ of CLASP, defined as:
\begin{equation}
    \mathcal{L}^{i}_{balancing}=\alpha [\text{CV}^{2}(\sum_{b=1}^{B}g_{b,e})+\text{CV}^{2}(\sum_{b=1}^{B} \mathbb{1}(g_{b,e}>0))].
\end{equation}
Here, $\alpha=0.01$ is the loss coefficient, $B$ is the batch size, and $g_{b,e}$ denotes the gating weight for the $e$-th expert assigned to the $b$-th sample, with $e \in \left\{1,\dots,N\right\}$. $\mathbb{1}$ is the indicator function. $\text{CV}^{2}(x)=\frac{\text{Var}(x)}{\text{Mean}(x)^2+\epsilon}$ (with $\epsilon=10^{-10}$) promotes balanced expert assignmentacross the batch.

We aggregate the load balancing losses from all stages to obtain the final load balancing loss, defined as: 
\begin{equation}
    \mathcal{L}_{balancing} = \sum_{i=1}^{N=4} \mathcal{L}^{i}_{balancing} .
\end{equation}
  
The total loss for training the CLASP framework is then formulated as:
\begin{equation}
\mathcal{L} = \lambda_{1} * \mathcal{L}_{dino} + \lambda_{2} * \mathcal{L}_{l} + \lambda_{3} *  \mathcal{L}_{h} + \lambda_{4} * \mathcal{L}_{balancing},
\end{equation}
where $\lambda_{1}$, $\lambda_{2}$, $\lambda_{3}$ and $\lambda_{4}$ are hyper-parameters that control the relative contribution of each loss component.
 \begin{figure*}[htb]
	\centering
	\setlength{\abovecaptionskip}{0.1cm}
	\includegraphics[width=1.00\textwidth]{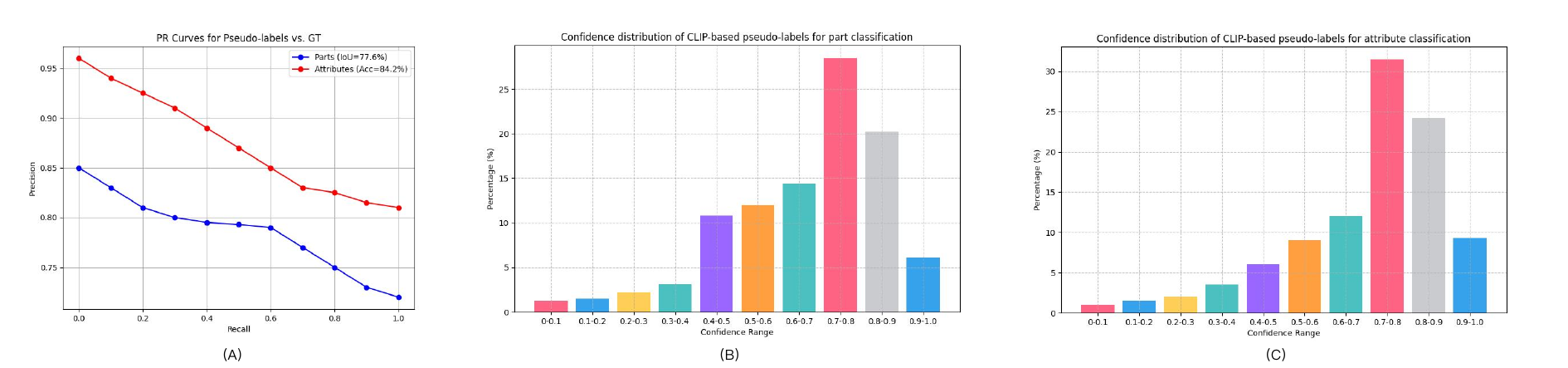}
	\caption{Precision-Recall curves and confidence scores statistics for CLIP-based part-level and attribute-level pseudo labels. (A) represents the PR(Precision-Recall) curves for CLIP-based parts/attributes pseudo labels on LIP and PA100k datasets. (B) represents the distribution of confidence scores for CLIP-based part pseudo-labels when $S=\{2,3,4\}$ on LUPerson. (C) represents the distribution of confidence scores for CLIP-based attribute pseudo-labels on LUPerson. }
	\label{pseudo_labels_statistics} 
    \vspace{-0.3cm}
\end{figure*}

 \begin{figure*}[t!]
	\centering
	\setlength{\abovecaptionskip}{0.1cm}
	\includegraphics[width=1.00\textwidth]{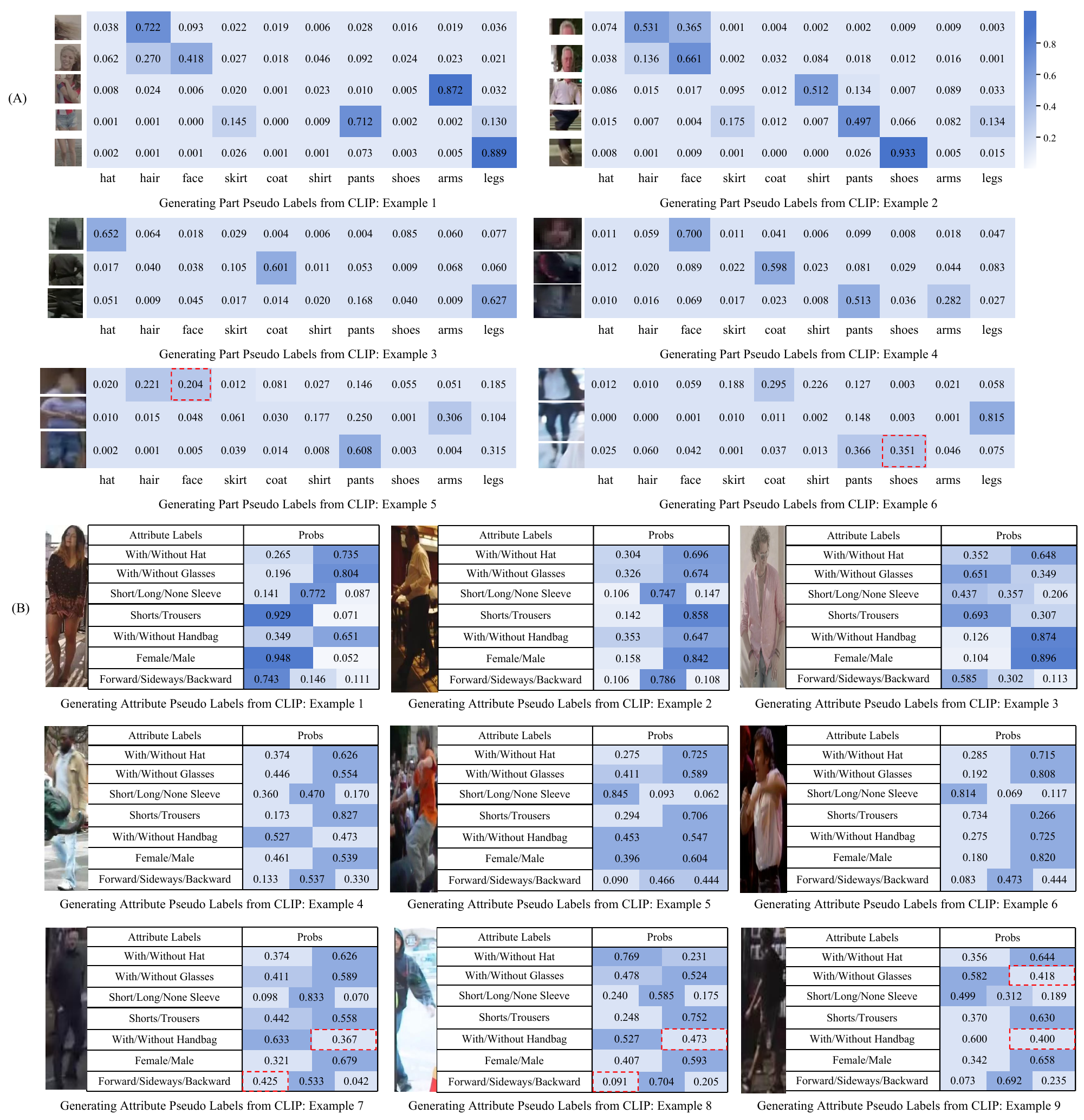}
	\caption{Examples of generating part-level and attribute-level pseudo labels using CLIP. (A) presents two examples where CLIP, combined with by predefined body-part texts, assign semantic labels and corresponding  probabilities to various clustered body parts. (B) presents three examples where CLIP, combined with predefined attribute texts, assign multiple attribute labels along with their confidence scores to different individuals. Red dashed boxes highlight ground-truth labels that were misclassified.}
	\label{pseudo_part_attribute_cases} 
    \vspace{-0.3cm}
\end{figure*}

\begin{figure*}
	\centering
	\setlength{\abovecaptionskip}{0.1cm}
	\includegraphics[width=1.00\textwidth]{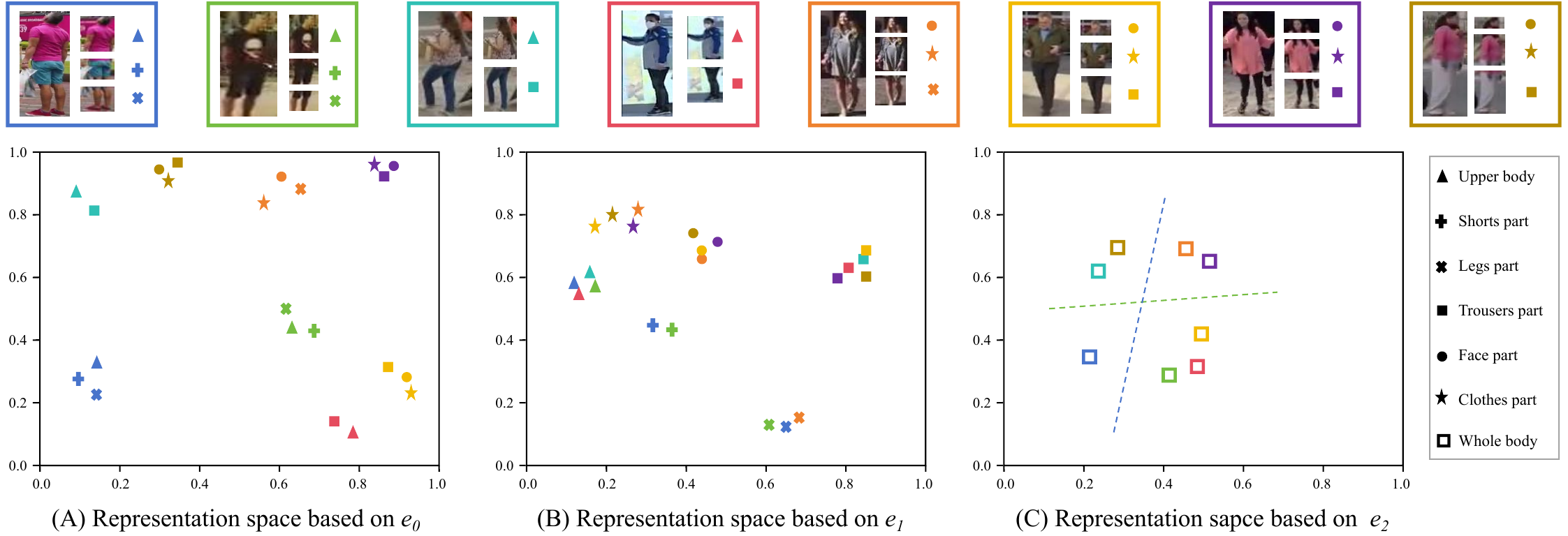}
	\caption{The representation space of part-level and whole-level feature embeddings under different task-specific prompts. Here, $e_0$ denotes the task prompt embedding used for contrastive learning in DINO, $e_1$ corresponds to the prompt for part-level classification, $e_2$ is the prompt for attribute-level classification.} \label{visualization_for_task_specific_embeddings} 
    \vspace{-0.5cm}
\end{figure*}

\section{Experiments}
In this section, we compare our proposed method against state-of-the-art approaches on human-centric downstream tasks, including person re-identification, attribute recognition, human parsing, person search, pose estimation and pedestrain detection, to demonstrate its generalization capability.

\vspace{-0.3cm}
\subsection{Experimental Setup}
\textbf{Datasets.} \
In this study, we adopt LUPerson \cite{fu2021unsupervised}, a large-scale dataset containing 4.18 million unlabeled pedestrian images, as our self-supervised pre-training data, consistent with the approach in SOLIDER \cite{chen2023beyond}. 
For downstream evaluation, we assess our method on a diverse set of human-centric tasks. For person reID, we evaluate on two widely-used datasets: Market-1501 \cite{zheng2015scalable} and MSMT17 \cite{wei2018person}. For attribute recognition, we conduct experiments on two datasets: PETA$_{zs}$\cite{jia2021rethinking} and PA100k \cite{liu2017hydraplus}.  For human parsing, we use the LIP dataset \cite{gong2017look} for evaluation. In addition, similar to SOLIDER, we also evaluate our proposed CLASP framework on three additional downstream tasks: person search  on CUHK-SYSU \cite{xiao2017joint} and PRW \cite{zheng2017person}, pedestrian detection on CityPerson \cite{zhang2017citypersons}, and pose estimation on COCO \cite{lin2014microsoft}.

\textbf{Evaluation Metrics.} \
Our evaluation protocol follows  SOLIDER \cite{chen2023beyond}. 
For the reID and person search tasks, we use mAP and Rank1 \cite{zheng2015scalable} as the evaluation metrics; For the attribute recognition task, we use mean accuracy (mA) \cite{jia2021rethinking} as the evaluation metric; In the human parsing task, we use mIoU \cite{li2020self} as the evaluation metric. We use the log-average Miss Rate over false positive on Reasonable and Heavy Occluded (MR$^{-2}$ on R/HO) to evaluate the performance on pedestrian detection task. For pose estimation, we utilize Average Precision/Recall (AP/AR) as the evaluation metric.

To concretely define and measure ``inter-task conflicts", we introduce three formal metrics:
\begin{itemize}
\item \emph{Gradient Conflict Ratio (GCR)}: We quantify the extent of gradient disagreement among tasks across shared network layers:
\begin{align}
\mathrm{CosSim}(g_i,g_j) &= \frac{\langle g_i,g_j\rangle}{\|g_i\|\,\|g_j\|}, \\
\mathrm{Conflict\ Ratio} &= \notag \\ \frac{1}{L}\sum_{\ell=1}^{L} \frac{1}{T(T-1)}
&\sum_{i\neq j} \mathbf{1}\big(\mathrm{CosSim}(g_i^{(\ell)},g_j^{(\ell)})<0\big),
\end{align}
where $g_i^{(\ell)}$ denotes the gradient of task $i$ at layer $l$, $T$ is the total number of tasks, and $L$ is the number of network layers. The ratio measures the fraction of task pairs exhibiting a negative cosine similarity between their gradients. 
This value reflects the frequency with which tasks drive shared parameters in conflicting directions. A higher GCR indicates a stronger degree of inter-task gradient conflict.

\item \emph{Expert-Activation Divergence (EAD)}: To analyze how tasks leverage experts within the PC-MoE module, we define the expert-activation divergence between task $i$ and task $j$ as:
\begin{align}
\textstyle D_{\mathrm{expert}}(i,j) = 1 - \frac{1}{N}\sum_{e=1}^{N} \min\!\big(p_e^{(i)}, p_e^{(j)}\big),
\end{align}
where $p_e^{(i)}$ and $p_e^{(j)}$ are the normalized activation probabilities of expert $e$ for task $i$ and $j$, respectively, and $N$ is the total number of experts. $D_{\mathrm{expert}}(i,j)$ measures the non-overlapping portion of expert activations. Larger EAD indicates that tasks tend to selectively activate distinct experts. This separation enables the network to decouple task-specific representations, thereby mitigating inter-task conflicts.

\item \emph{Harmonic Mean (HM) of Task Performance}: To provide a balanced assessment of  overall multi-task effectiveness, we compute the harmonic mean of the key metrics for person ReID and semantic parsing:
\begin{align}
\textstyle \mathrm{HM(ReID, Parsing)} = \frac{2 \cdot \mathrm{mAP_{ReID}} \cdot \mathrm{mIoU_{Parsing}}}{\mathrm{mAP_{ReID}} + \mathrm{mIoU_{Parsing}}}.
\end{align}
The metric highlights balanced improvements, as it severely penalizes scenarios where one task performs well while the other lags. Consequently, it offers a robust measure of the integrated performance across the reID and parsing tasks.
\end{itemize}
 
\textbf{Implementation Details.} \
All experiments are conducted on 8 Ascend 910B4 32G GPUs, with batch sizes set to $96/64/36$ for swin-tiny/swin-small/swin-base, respectively. We employ AdamW \cite{loshchilov2017decoupled} as the optimizer with an initial learning rate of 0.0005, and adopt a cosine annealing scheduler for learning rate adjustment. 
Due to the computational overhead of semantic clustering, we adopt a two-stage training strategy to accelerate pre-training.  Specifically, we first train a base DINO framework, which serves as our baseline, and subsequently fine-tune it with our proposed CLASP framework for 30 epochs using a reduced learning rate of 0.0003. For semantic label generation, we adopt ViT-B/32 as the vision-language model to extract part-level and attribute-level semantics.  In PC-MoE module, we set the number of expert networks $N=10$ and the number of selected experts $K=6$. The loss weights are set as follows: $\lambda_1=0.8$, $\lambda_2=0.6$, $\lambda_3=0.6$ and $\lambda_4=1.0$. 

During fine-tuning, the CLASP-pretrained backbone is integrated into task-specific model frameworks. Meanwhile, we denote the task-specific prompt vectors corresponding to the three pre-training tasks in CLASP (DINO-based contrastive learning, part-level classification, and attribute-level classification) as \(e_0, e_1\) and \(e_2\), respectively. These prompt vectors guide the model to produce features that emphasize appearance, part-level semantics, and attribute-level semantics, correspondingly. To accommodate the characteristics of different downstream tasks, we employ distinct task-specific prompt vectors during the fine-tuning stage. Following SOLIDER, for person ReID, we use TransReID \cite{he2021transreid} for fine-tuning, guided by the prompt vector \( e_0 \).  For human parsing, we employ SCHP \cite{li2020self}, guided by the prompt vector \( e_1 \).  For pedestrian detection, we adopt CSP \cite{liu2019high} with prompt vector \( e_1 \). For pose estimation, we utilize HRFormer \cite{yuan2021hrformer}, also guided by prompt vector \( e_1 \), as both tasks benefit from part-aware semantic representations. For person search, we fine-tune using SeqNet \cite{li2021sequential}, leveraging both prompt vectors \( e_0 \) and \( e_1 \) to jointly model appearance and part-level cues.  For attribute recognition, we adopt RethinkPAR \cite{jia2021rethinking}, incorporating the prompt vector \( e_2 \). 
\vspace{-0.3cm}
\subsection{Qualitative and Quantitative Analysis}

\textbf{Analysis of the reliability of CLIP-based Pseudo Labels.} \
To assess the reliability of pseudo-labels generated by CLIP-based pipeline, we conduct quantitative evaluations on two human-annotated benchmarks: LIP for human part segmentation and PA100k for human attribute recognition. (i) For LIP, which primarily contains full-body human images, the number of token clusters is fixed at 4. Reliability is measured by the mean Intersection over Union (mIoU) between each predicted part region and its ground-truth counterpart, averaged across all semantic parts. (ii) For PA100k, CLIP-derived attribute pseudo-labels are evaluated in term of classification accuracy with respect to ground-truth annotations.
As shown in Fig.~\ref{pseudo_labels_statistics} (A), the part pseudo-labels achieve competitive performance, with mIoU values ranging from $72\%$ to $85\%$ and an average of  $78\%$. The attribute pseudo-labels achieve an average classification accuracy of $84.2\%$.  Collectively, these quantitative results emphatically demonstrate that the CLIP-generated pseudo-labels are quantitatively reliable and thus suitable to serve as effective supervisory signals for subsequent self-supervised pretraining. Moreover, as shown in Fig.~\ref{pseudo_labels_statistics} (B) and (C), the confidence distributions on LUPerson indicate strong reliability, with over 81.2\% of samples exceeding a confidence score of 0.5 for part pseudo-labels when $S=\{2,3,4\}$, and over 65\% exceeding 0.7 for attribute pseudo-labels.
These statistics collectively demonstrate that the CLIP-based part and attribute pseudo-labels are reliable and semantically well-grounded.

\textbf{Analysis of Pseudo Part-level Labels Generated by CLIP.} \ 
To assess CLIP's ability to classify semantic body parts, we first crop the clustered regions and process them using the CLIP image encoder. These image embeddings are then compared against text embeddings generated by CLIP text encoder, assigning pseudo labels based on the highest similarity scores. Fig.~\ref{pseudo_part_attribute_cases} (A) presents two examples of CLIP-generated part-level pseudo labels. The accompanying tables present similarity-based probability scores between each image region and a set of predefined part-level text labels. The label with the highest score (highlighted in red) is selected as the final pseudo label. As shown in Fig.~\ref{pseudo_part_attribute_cases} (A), in Example 1, the top-scoring labels are: \textit{hair} (0.722),  \textit{face} (0.418), \textit{arm} (0.872), \textit{pants} (0.712) and \textit{legs} (0.889); In Example 2, the top-scoring labels are: \textit{hair} (0.531), \textit{face} (0.661), \textit{shirt} (0.512), \textit{pants} (0.497) and \textit{pants} (0.933). Examples 3 and 4 demonstrate that CLIP can achieve accurate part classification even in cases of back-facing views and partial occlusion.
These results demonstrate the reliability and discriminative capability of CLIP in assigning meaningful pseudo part-level labels, validating its utility for weakly supervised semantic supervision. Meanwhile, as shown in Example 5 and Example 6, we also observe that when certain body parts are blurry or degraded, the part semantic labels predicted by CLIP can be misclassified. Crucially, the proposed quality filtering process in Sec.~\ref{sec:3.1} actively maintains input quality by removing low-quality images, thereby mitigating these issues.

\textbf{Analysis of Pseudo Attribute-level Labels Generated by CLIP.} \
We evaluate CLIP's capability in generating attribute-level pseudo labels. Fig.~\ref{pseudo_part_attribute_cases} (B) presents three representative examples, each  accompanied by a table displaying similarity-based probability scores for various attributes.
The attribute with the highest score (highlighted in red) is selected as the pseudo label.  In Example 1, the top-scoring attributes are: \textit{Without Hat} (0.735), \textit{Long Sleeve} (0.772), and \textit{Female} (0.948). 
In Example 2, the highest scores are for: \textit{Trousers} (0.858), \textit{Male} (0.842), and \textit{Sideways} (0.786).
In Example 3, the most probable labels include: \textit{With Glasses} (0.651), \textit{Shorts} (0.693), and \textit{Male} (0.896). Examples 4, 5, and 6 demonstrate that CLIP can achieve accurate part classification under various hard cases, such as partial occlusion, multiple people, and local blurriness.
These results demonstrate CLIP’s strong alignment between visual content and textual semantics, confirming its effectiveness in producing reliable and accurate attribute-level pseudo labels that are well-aligned with the actual visual attributes in the images. Additionally, as shown in Examples 3-6, LIP-generated attribute pseudo-labels are prone to errors in cases with limited human region visibility, color similarity between the human and background, or excessive blurriness. We also apply the proposed quality filtering process in Sec.~\ref{sec:3.1} to filter these low-quality human images.

\textbf{Analysis of Task-specific Prompt Embeddings.} \
To analyze the representation spaces of part-level and whole-level feature embeddings under different task-specific prompt embeddings, we present the visualizations in Fig.~\ref{visualization_for_task_specific_embeddings}. The task prompt \( e_0 \), designed for contrastive learning in DINO, focuses on global image characteristics, enabling effective  discrimination  between individuals. As shown, feature representations from different parts of the same person are highly similar, while those from the same part across different individuals exhibit clear separation, reflecting the global identity-centric nature of the task.
Fig.~\ref{visualization_for_task_specific_embeddings} (B) illustrates the effect of prompt \( e_1 \), which is tailored for part-level classification. Here,
the model captures localized semantics, forming distinct clusters for specific body parts such as the upper body, face, trousers, and clothing regions.
In Fig.~\ref{visualization_for_task_specific_embeddings} (C), using prompt \( e_2 \),
for attribute-level classification, the visualization reveals higher-level semantic organization. The green dashed line separates feature vectors based on gender, while the blue dashed line distinguishes features by viewpoint/orientation. These results collectively demonstrate the flexibility and task-adaptability of our framework, which can shift between global contrastive learning and fine-grained part or attribute-level representation learning by leveraging task-specific prompt embeddings.

\begin{table*}[htp]
\centering
\caption{Ablation study on pseudo part-level and attribute-level semantic labels generated by CLIP. In the \textbf{Single-granularity (SG)} setting, k-means uses 3 clusters (excluding background); the \textbf{Multi-granularity (MG)} setting randomly selects from 2,3, or 4 clusters at each pre-training iteration. \textbf{Pseudo-Part-Labels (PPL)} and \textbf{Pseudo-Attribute-Labels (PAL)} replace clustering-based labels with CLIP-derived part-level and attribute-level labels, respectively. $\uparrow$/$\downarrow$ indicate that higher/lower values is better.}
\vspace{-0.15cm}
\label{ablation_study_clip_pseudo_labels}
\resizebox{\textwidth}{!}{
    \begin{tabular}{@{}cccccccccc@{}}
    \toprule
    \multirow{2}{*}{Model Number} & \multicolumn{4}{c}{Methods} & \multicolumn{2}{c}{Person Re-identification (mAP/Rank1 $\uparrow$)} & Attribute Recognition (mA $\uparrow$) & Human Parsing (mIoU $\uparrow$) \\ 
    \cmidrule(lr){2-5} \cmidrule(lr){6-7} \cmidrule(lr){8-8} \cmidrule(lr){9-9}
    &  SG & MG & PPL & PAL & Market1501 & MSMT17 & PA100k & LIP \\
    \midrule
    $\mathrm{A}$ & \multicolumn{4}{c}{Baseline (the standard DINO \cite{caron2021emerging})} & \textbf{88.5/95.2} & \textbf{62.7/83.0} & 82.82 & 55.13 \\ \hline \specialrule{0em}{1pt}{1pt}
    $\mathrm{B}$ & \checkmark &  &  &  & 88.3/95.1 & 61.9/82.3 & 82.80 & 55.25 \\ 
    $\mathrm{C}$ &  & \checkmark &  &  & 88.0/95.1 & 61.5/82.2 & 82.66 & 55.49 \\ 
    $\mathrm{D}$ &  & \checkmark & \checkmark &  & 87.7/95.0 & 61.2/81.9 & 82.47 & \textbf{55.78} \\ 
    $\mathrm{E}$ &  &  &  & \checkmark & 87.9/95.1 & 61.0/81.7 & \textbf{83.42} & 55.02 \\ 
    \midrule
    $\mathrm{F}$ &  & \checkmark & \checkmark & \checkmark & 87.4/94.6 & 60.5/81.2 & 82.86 & 55.31 \\ 
    \bottomrule
    \end{tabular}}
\vspace{-0.3cm}
\end{table*} 

\begin{table*}
\caption{Ablation study on the PC-MoE module. \textbf{Self-embeddings (SE)} use the output of stage $i$ as input to the gating network, while \textbf{Prompt-embeddings (PE)} use task-specific learnable prompts. \textbf{Channel-gate (CG)} assigns weights to individual channels, and \textbf{Global-gate (GG)} assigns a single weight to each expert's output. \textbf{Prompt-Res (PR)} transforms the prompt via a fully connected layer and integrates it into the MoE output in a residual manner.}
\vspace{-0.15cm}
\label{ablation_study_prompt_controlled_moe}
\resizebox{\textwidth}{!}{
    \begin{tabular}{@{}ccccccccccc@{}}
    \toprule
    \multirow{2}{*}{Model Number} & \multicolumn{5}{c}{Methods} & \multicolumn{2}{c}{Person Re-identification (mAP/Rank1 $\uparrow$)} & Attribute Recognition (mA $\uparrow$) & Human Parsing (mIoU $\uparrow$) \\ 
    \cmidrule(lr){2-6} \cmidrule(lr){7-8} \cmidrule(lr){9-9} \cmidrule(lr){10-10}
    &  SE & PE & CG & GG & PR & Market1501 & MSMT17 & PA100k & LIP \\
    \midrule
    $\mathrm{F}$ & \multicolumn{5}{c}{Baseline(F method in Tab.~\ref{ablation_study_clip_pseudo_labels})} & 87.4/94.6 & 60.5/81.2 & 82.86 & 55.31\\ \hline \specialrule{0em}{1pt}{1pt}
    $\mathrm{G}$ & \checkmark &  &  & \checkmark & &  88.3/95.2 & 62.4/82.7 & 83.27 & 55.45  \\ 
    $\mathrm{H}$ &  & \checkmark &  & \checkmark & & 88.5/95.4 & 62.6/82.8 & 83.51 & 55.70 \\ 
    $\mathrm{I}$ & \checkmark & \checkmark &  & \checkmark & & 88.8/95.4 & 62.9/83.1 & 83.77 & 55.92 \\ 
    $\mathrm{J}$ & \checkmark & \checkmark & \checkmark & & & 88.6/95.2 & 62.7/82.7 & 83.58 & 55.64 \\ 
    $\mathrm{K}$ & \checkmark & \checkmark & \checkmark & \checkmark  & & 89.1/95.5 & 63.3/83.5 & 84.02 & 56.17 \\ 
    \midrule
    $\mathrm{L}$ & \checkmark & \checkmark & \checkmark & \checkmark & \checkmark & \textbf{89.2/95.7} & \textbf{63.8/83.5} & \textbf{84.26} & \textbf{56.39} \\ 
    \bottomrule
    \end{tabular}}
\vspace{-0.3cm}
\end{table*}

\begin{table*}
\caption{Ablation study on MoE design and usage. \textbf{One-gate (OG)} uses a shared gating network for all tasks, while \textbf{Multi-gates (MG)} assigns one per pre-training task. \textbf{PC-gate (PCG)} incorporates task-specific prompt vectors to guide expert selection. \textbf{Last-stage (LS)} applies MoE only to the final-stage DINO features, whereas \textbf{All-stages (AS)} extends it to all stages for finer-grained control.}
\vspace{-0.15cm}
\label{ablation_study_different_gates_stages}
\resizebox{\textwidth}{!}{
    \begin{tabular}{@{}ccccccccccc@{}}
    \toprule
    \multirow{2}{*}{Model Number} & \multicolumn{5}{c}{Methods} & \multicolumn{2}{c}{Person Re-identification (mAP/Rank1 $\uparrow$)} & Attribute Recognition (mA $\uparrow$) & Human Parsing (mIoU $\uparrow$) \\ 
    \cmidrule(lr){2-6} \cmidrule(lr){7-8} \cmidrule(lr){9-9} \cmidrule(lr){10-10}
    & OG & MG & PCG & LS & AS & Market1501 & MSMT17 & PA100k & LIP \\
    \midrule
    $\mathrm{F}$ & \multicolumn{5}{c}{Baseline(F method in Tab.~\ref{ablation_study_clip_pseudo_labels})} & 87.4/94.6 & 60.5/81.2 & 82.86 & 55.31\\ \hline \specialrule{0em}{1pt}{1pt}
    $\mathrm{G}$ & \checkmark &  &  & \checkmark &  & 88.3/95.2 & 62.4/82.7 & 83.27 & 55.45  \\ 
    $\mathrm{L}$ &  &  & \checkmark & \checkmark &  & 89.2/95.7 & 63.5/83.5 & 84.26 & 56.39 \\ 
    $\mathrm{M}$ &  & \checkmark &  & \checkmark &  & 88.6/95.3 & 62.6/83.0 & 83.88 & 56.06 \\ 
    \midrule
    $\mathrm{N}$ & \checkmark &  &  &  & \checkmark & 88.6/95.4 & 62.8/83.3 & 83.74 & 55.91\\ 
    $\mathrm{O}$ &  & \checkmark &  &  & \checkmark & 89.3/95.7 & 63.0/83.5 & 84.22 & 56.53\\ 
    $\mathrm{P}$ &  &  & \checkmark &  & \checkmark & \textbf{89.7/96.2} & \textbf{63.6/83.9} & \textbf{84.86} & \textbf{57.02} \\ 
    \bottomrule
    \end{tabular}}
\vspace{-0.2cm}
\end{table*}

\begin{table*}
\centering
\scriptsize
\caption{Sensitivity analysis on $S$ (the cluster number set), $N$ (the number of expert networks), $M$ (the number of task-specific prompt vectors), and Top-$K$ (routing).}
\label{sensitivity_for_knmtopk}
\resizebox{\textwidth}{!}{
\begin{tabular}{@{}llccccc@{}}
\toprule
Method & $S$ (cluster number set) & $N$ (expert networks) & Top-$K$ (activation number) & $M$ (task prompt vectors) & ReID (mAP/Rank1 $\uparrow$) & Parsing (mIoU $\uparrow$) \\
\midrule
Default & $S_4=\{2,3,4\}$ & 10 & 6 & 3 & 94.6/97.2 & 61.74 \\
Var-1   & $S_3=\{2,3\}$ & 10 & 6 & 3 & 94.3/97.0 & 61.42 \\
Var-2   & $S_5=\{2,3,4,5\}$ & 10 & 6 & 3 & 94.5/97.1 & 61.61 \\
Var-3   & $S_4=\{2,3,4\}$ & 6  & 6 & 3 & 94.0/96.7 & 61.18 \\
Var-4   & $S_4=\{2,3,4\}$ & 12 & 6 & 3 & 94.7/97.3 & 61.85 \\
Var-5   & $S_4=\{2,3,4\}$ & 10 & 4 & 3 & 94.2/96.9 & 61.35 \\
Var-6   & $S_4=\{2,3,4\}$ & 10 & 8 & 3 & 94.7/97.3 & 61.81 \\
Var-7   & $S_4=\{2,3,4\}$ & 10 & 6 & 2 & 94.1/96.8 & 61.22 \\
Var-8   & $S_4=\{2,3,4\}$ & 10 & 6 & 6 & 94.6/97.2 & 61.79 \\
\bottomrule
\end{tabular}}
\label{tab1}
\end{table*}

\subsection{Ablation Study}
In this section, we conduct an ablation study of our proposed CLASP framework on three  human-centric downstream tasks: person reID, attribute recognition, and human parsing.
we randomly select 1 million images from the LUPerson dataset for pre-training and  perform the experiments using the Swin-Tiny backbone. 

\textbf{The Effectiveness of Pseudo Part-level Labels.} \ 
To evaluate the effectiveness of part-level semantic labels generated by CLIP, we design several experimental settings. As shown in Tab.~\ref{ablation_study_clip_pseudo_labels}, adopting multi-granularity (MG) clustering (randomly selecting the number of clusters during training) yields an improvement in human parsing performance (+0.24\% mIoU) compared to single-granularity (SG) clustering. However, this also results in a slight drop in ReID performance, likely due to shifts in the feature distribution.
When CLIP-generated pseudo part labels are incorporated into the multi-granularity clustering setting, human parsing performance is further enhanced (+0.65\% mIoU), indicating that part-level semantic supervision effectively guides part-aware feature learning. Nevertheless, this benefit comes at the cost of ReID performance, as the increased focus on localized features reduces the model's ability to capture global appearance cues, which are essential for person ReID.
These findings highlight a key trade-off: while pseudo part-level labels (PPL) improve fine-grained tasks such as human parsing, they may negatively impact tasks like ReID that rely more on global discriminative features.

\textbf{The Effectiveness of Pseudo Attribute-level Labels.} \
We further investigate the impact of pseudo attribute-level labels (PAL) generated by CLIP for self-supervised pre-training. By integrating pseudo attribute labels into the DINO framework, we aim to enhance attribute recognition performance. As shown in Tab.~ \ref{ablation_study_clip_pseudo_labels}, incorporating CLIP-guided attribute labels (Model E) leads to an improvement in attribute recognition, with the mean accuracy (mA) increasing from 82.82\% to 83.42\%. However, this improvement comes with a slight performance trade-off on other tasks: ReID performance drops modestly from 62.7\% to 61.0\% mAP on MSMT17, and Human parsing performance decreases slightly from 55.13\% to 55.02\% mIoU. 
 These results suggest that while the pseudo attribute labels are beneficial for improving attribute recognition, they may reduce model's effectiveness in tasks like ReID and human parsing.

\textbf{The Effectiveness of Prompt-Controlled MoE Module.} \ 
Tab.~\ref{ablation_study_prompt_controlled_moe} conducts an ablation study to evaluate the contribution of different components within the prompt-controlled MoE (PC-MoE) module. We can observe that: \textbf{(1)} Introducing self-embeddings (SE) and a global-gate (GG) yields improvements in reID, with gains of 0.9\% mAP on Market1501 and 1.9\% mAP on MSMT17, while also slightly enhancing attribute recognition and human parsing performance. 
\textbf{(2)} Replacing self-embeddings with prompt-embeddings (PE) leads to further improvements, with a 0.24\% mA increase for attribute recognition and a 0.25\% mIoU gain for human parsing. Combining both self- and prompt embeddings yields additional gains across all tasks, demonstrating their complementary roles in adapting to different semantic requirements.
\textbf{(3)} Incorporating a channel gate (CG) boosts attribute recognition performance, though its effect on ReID is relatively modest. When both channel gate and global gate are applied, we observe more consistent improvements across tasks, including a 1.7\% mAP gain on Market1501 and a 1.16\% mA gain in attribute recognition.
\textbf{(4)} Finally, integrating prompt-Res (PR), our residual prompt enhancement module, achieves the best overall performance, with improvements of 1.8\% mAP on Market1501, 3.3\% mAP on MSMT17, 1.4\% mA in attribute recognition, and 1.08\% mIoU in human parsing. These results demonstrate the effectiveness and flexibility of PC-MoE, particularly when enhanced with prompt-Res, in adapting to diverse human-centric visual tasks.

\textbf{The Effectiveness of Different MoE Designs and Their Usage.} \
Tab.~ \ref{ablation_study_different_gates_stages} conducts an ablation study to evaluate the effect of different MoE designs and their usage strategies. We can observe that: \textbf{(1)} Using a single gate (One-gate, OG) at the final stage improves  0.9\% mAP on Market1501 and 1.9\% mAP on MSMT17 compared to the baseline. And introducing Multi-gates (MG) yields further improvements of 1.2\% on Market1501 and 2.1\% on MSMT17, indicating the benefit of more granular expert control. 
\textbf{(2)} Comparing method L with G, we observe that incorporating the Prompt-Controlled Gate (PC-gate) significantly enhances performance across all tasks:
0.9\% mAP on Market1501, 1.1\% mAP on MSMT17, 0.99\% mA for attribute recognition, and 0.94\% mIoU for human parsing. Further applying gating across all transformer stages (All-stages, AS) consistently boosts performance regardless of the gating design. These findings clearly demonstrate the effectiveness and scalability of the proposed CLASP framework, and highlight the value of prompt-controlled, multi-stage expert gating in addressing diverse human-centric visual tasks.

\textbf{The Effectiveness of different hyperparameters (number of clusters, task-specific learnable prompts, expert networks, and Top-K activation).} \
To validate hyperparameter robustness, as shown in Tab.~\ref{sensitivity_for_knmtopk}, we conducted ablation experiments on four key settings. For the multi-granularity K-means cluster number ($L$), $L$ is sampled from a set $S$, enabling adaptation to pedestrian image variations (e.g., partial vs. full-body visibility). Evaluations on $S_3=\{2,3\}$, $S_4=\{2,3,4\}$, $S_5=\{2,3,4,5\}$ via ReID and human parsing tasks show only minor performance fluctuations, confirming robustness. For task-specific learnable prompts ($M$), Tab.~\ref{sensitivity_for_knmtopk} indicates $M=2$ underperforms due to insufficient feature disentanglement, while $M=6$ yields marginal gains but higher computational cost, leading to $M=3$ being chosen as the default. For the number of expert networks ($N$) and Top-K activation, the results show performance variations of $<1\%$ across different settings of both $N$ and the Top-K candidate count, verifying the method’s robustness to the expert pool scale and the selection criterion for activated candidates.

\begin{table}[t!]
\small
\tabcolsep=2pt
\begin{minipage}[t]{0.45\textwidth}
\centering
\caption{Comparison with state-of-the-art methods on ReID. CLASP (TransReID) refers to TransReID with the pre-trained backbone based on CLASP.}
\vspace{-0.15cm}
\label{tab:reid}
\begin{threeparttable}
\begin{tabular}{llcccc}
\hline
\specialrule{0em}{1pt}{1pt}
\multirow{2}{*}{\textbf{Method}} & \multirow{2}{*}{\textbf{Backbone}} & \multicolumn{2}{c}{\textbf{Market1501}}  & \multicolumn{2}{c}{\textbf{MSMT17}}  \\ \cmidrule(lr){3-4} \cmidrule(lr){5-6} &  & \textbf{mAP}$\uparrow$ & \textbf{Rank1}$\uparrow$ & \textbf{mAP}$\uparrow$ & \textbf{Rank1}$\uparrow$ \\ 
       \hline
\specialrule{0em}{1pt}{1pt}
TransReID~\cite{he2021transreid} & ViT-B & 89.5 & 95.2 & 69.4 & 86.2 \\
UP-ReID~\cite{yang2022unleashing} & Res-50 & 91.1 & 97.1 & 63.3 & 84.3 \\
PASS~\cite{zhu2022pass} & ViT-B & 93.3 & 96.9 & 74.3 & 89.7 \\
SOLIDER~\cite{chen2023beyond} & Swin-B & 93.9 & 96.9 & 77.1 & 90.7 \\ 
UniHCP~\cite{ci2023unihcp} & ViT-B & 90.3 & - & 67.3 & - \\
PATH~\cite{tang2023humanbench} & ViT-B & 89.5 & - & 69.1 & - \\
HAP~\cite{yuan2023hap} & ViT-B & 93.9 & - & 78.1 & - \\ 
PersonMAE~\cite{hu2024personmae} & ViT-B & 93.6 & \textbf{97.1} & \textbf{79.8} & \underline{91.4}  \\
\hline
\specialrule{0em}{1pt}{1pt}
CLASP (TransReID) & Swin-T & 92.4 & 96.3 & 68.1 & 86.8 \\
CLASP (TransReID) & Swin-S & \underline{94.3} & 96.7 & 77.2 & 91.2 \\
CLASP (TransReID) & Swin-B & \textbf{94.6} & \textbf{97.1} & \underline{78.8} & \textbf{91.8} \\
\specialrule{0em}{1pt}{1pt} 
\hline
\end{tabular}
\end{threeparttable}
\end{minipage}

\begin{minipage}{\columnwidth}
\centering
\caption{Comparison with state-of-the-art methods on Attribute Recognition. CLASP (RethinkPAR) refers to RethinkPAR with the pre-trained backbone based on CLASP.}
\vspace{-0.15cm}
\label{tab:attribute_recog}
\begin{threeparttable}
\begin{tabular}{llccc}
\hline
\specialrule{0em}{1pt}{1pt}
\multirow{2}{*}{\textbf{Method}} & \multirow{2}{*}{\textbf{Backbone}} & \multicolumn{1}{c}{\textbf{PETA$_{ZS}$}}  & \multicolumn{1}{c}{\textbf{RAP$_{ZS}$}} & \multicolumn{1}{c}{\textbf{PA100k}}  \\ \cmidrule(lr){3-3} \cmidrule(lr){4-4} \cmidrule(lr){5-5}  &  & \textbf{mA}$\uparrow$ & \textbf{mA}$\uparrow$ & \textbf{mA}$\uparrow$ \\ 
       \hline
\specialrule{0em}{1pt}{1pt}
MsVAA~\cite{sarafianos2018deep} & Res-101 & 71.53 & 72.04 & 80.41  \\
VAC~\cite{guo2019visual} &  Res-101 & 71.91 & 73.70 & 79.16  \\
RethinkPAR~\cite{jia2021rethinking} & - & 71.62 & 72.32 & 81.61 \\
SOLIDER~\cite{chen2023beyond} & Swin-B & 76.43 & 77.06 & 86.37  \\ 
UniHCP~\cite{ci2023unihcp} & ViT-B & - & - & 86.18 \\
PATH~\cite{tang2023humanbench} & ViT-B & - & - & 85.0 \\
HAP~\cite{yuan2023hap} & ViT-B & - & - & 86.54 \\
Hulk~\cite{wang2023hulk} & ViT-B & - & - & 82.85 \\
\hline
\specialrule{0em}{1pt}{1pt}
CLASP (RethinkPAR) & Swin-T & 76.03 & 75.66 & 85.32 \\
CLASP (RethinkPAR) & Swin-S & \underline{76.64} & \underline{77.39} & \underline{86.88}  \\
CLASP (RethinkPAR) & Swin-B & \textbf{77.86} & \textbf{77.82} & \textbf{87.91}  \\
\specialrule{0em}{1pt}{1pt} 
\hline
\end{tabular}
\end{threeparttable}
\end{minipage}

\begin{minipage}{\columnwidth}
\centering
\caption{Comparison with state-of-the-art methods on Person Search. CLASP (SeqNet) refers to SeqNet with the pre-trained backbone based on CLASP.}
\vspace{-0.15cm}
\label{tab:person_search}
\begin{threeparttable}
\begin{tabular}{llcccc}
\hline
\specialrule{0em}{1pt}{1pt}
\multirow{2}{*}{\textbf{Method}} & \multirow{2}{*}{\textbf{Backbone}} & \multicolumn{2}{c}{\textbf{CUHK-SYSU}}  & \multicolumn{2}{c}{\textbf{PRW}}  \\ \cmidrule(lr){3-4} \cmidrule(lr){5-6} &  & \textbf{mAP}$\uparrow$ & \textbf{Rank1}$\uparrow$ & \textbf{mAP}$\uparrow$ & \textbf{Rank1}$\uparrow$ \\ 
       \hline
\specialrule{0em}{1pt}{1pt}
TCTS~\cite{wang2020tcts} & Res-50  & 93.9 & 95.1 & 46.8 & 87.5 \\
SeqNet~\cite{li2021sequential} & Res-50 & 94.8 & 95.7 & 47.6 & 87.6 \\
GLCNet~\cite{qin2023movienet} & Res-50 & \underline{95.8} & \underline{96.2} & 47.8 & \textbf{87.8} \\
SOLIDER~\cite{chen2023beyond} & Swin-B & 94.9 & 95.5 & 59.7 & 86.8 \\ 
\hline
\specialrule{0em}{1pt}{1pt}
CLASP (SeqNet) & Swin-T & 95.1 & 95.5 & 57.3 & 87.1 \\
CLASP (SeqNet) & Swin-S & 95.7 & 95.9 & \underline{60.6} & 87.4 \\
CLASP (SeqNet) & Swin-B & \textbf{96.1} & \textbf{96.4} & \textbf{60.8} & \textbf{87.8} \\
\specialrule{0em}{1pt}{1pt} 
\hline
\end{tabular}
\end{threeparttable}
\end{minipage}
\vspace{-0.4cm}
\end{table}

\begin{table}[t!]
\small
\tabcolsep=2pt
\begin{minipage}{\columnwidth}
\centering
\caption{Comparison with state-of-the-art methods on Human Parsing. CLASP (SCHP) refers to SCHP with the pre-trained backbone based on CLASP.} \vspace{-0.15cm}
\label{tab:human_parsing}
\begin{threeparttable}
\begin{tabular}{llc}
\hline
\specialrule{0em}{1pt}{1pt}
\multirow{2}{*}{\textbf{Method}} & \multirow{2}{*}{\textbf{Backbone}} & \multicolumn{1}{c}{\textbf{LIP}}  \\ \cmidrule(lr){3-3}  &  & \textbf{mA}$\uparrow$ \\ 
       \hline
\specialrule{0em}{1pt}{1pt}
CE2P~\cite{ruan2019devil} & Res-101 & 53.10  \\
PCNet~\cite{zhang2020part} & Res-101 & 57.03  \\
SCHP~\cite{li2020self} & Res-101 & 59.36 \\
SOLIDER~\cite{chen2023beyond} & Swin-B & 60.50   \\ 
Hulk~\cite{wang2023hulk} & ViT-B & \textbf{63.95} \\
\hline
\specialrule{0em}{1pt}{1pt}
CLASP (SCHP) & Swin-T & 59.31  \\
CLASP (SCHP) & Swin-S & 60.93  \\
CLASP (SCHP) & Swin-B & \underline{61.74}   \\
\specialrule{0em}{1pt}{1pt} 
\hline
\end{tabular}
\end{threeparttable}
\end{minipage}

\begin{minipage}{\columnwidth}
\centering
\caption{Comparison with state-of-the-art methods on Pose Estimation. CLASP (HRFormer) refers to HRFormer with the pre-trained backbone based on CLASP.} \vspace{-0.15cm}
\label{tab:pose_estimation}
\begin{threeparttable}
\begin{tabular}{llcc}
\hline
\specialrule{0em}{1pt}{1pt}
\multirow{2}{*}{\textbf{Method}} & \multirow{2}{*}{\textbf{Backbone}} & \multicolumn{2}{c}{\textbf{COCO}}  \\ \cmidrule(lr){3-4}  &  & \textbf{AP}$\uparrow$ & \textbf{AR}$\uparrow$ \\ 
       \hline
\specialrule{0em}{1pt}{1pt}
HRNet \cite{sun2019deep} & HRNet-W48 & 76.3 & 81.2 \\
HRFormer \cite{yuan2021hrformer} & HRFormer-B & 77.2 & \underline{82.0}  \\
SOLIDER~\cite{chen2023beyond} & Swin-B & 76.6 & 81.5 \\ 
UniHCP~\cite{ci2023unihcp} & ViT-B & 76.5 & -  \\
PATH~\cite{tang2023humanbench} & ViT-B & 76.3 & -  \\
HAP~\cite{yuan2023hap} & ViT-B & \textbf{78.2} & -  \\ 
Hulk~\cite{wang2023hulk} & ViT-B & 77.0 & - \\
\hline
\specialrule{0em}{1pt}{1pt}
CLASP (HRFormer) & Swin-T & 75.7 & 80.3  \\
CLASP (HRFormer) & Swin-S & 76.8 & 81.8 \\
CLASP (HRFormer) & Swin-B & \underline{77.8} & \textbf{82.2}   \\
\specialrule{0em}{1pt}{1pt} 
\hline
\end{tabular}
\end{threeparttable}
\end{minipage}

\begin{minipage}{\columnwidth}
\centering
\caption{Comparison with state-of-the-art methods on Pedestrian Detection. CLASP (CSP) refers to CSP with the pre-trained backbone based on CLASP.} \vspace{-0.15cm}
\label{tab:pedestrian_detection}
\begin{threeparttable}
\begin{tabular}{llcc}
\hline
\specialrule{0em}{1pt}{1pt}
\multirow{2}{*}{\textbf{Method}} & \multirow{2}{*}{\textbf{Backbone}} & \multicolumn{2}{c}{\textbf{CityPerson}}  \\ \cmidrule(lr){3-4} &  & \textbf{ R (MR$^{-2}$)}$\downarrow$ & \textbf{HO (MR$^{-2}$)}$\downarrow$ \\ 
       \hline
\specialrule{0em}{1pt}{1pt}
RepLoss \cite{wang2018repulsion} & - & 13.2 & 56.9 \\
CSP \cite{liu2019high} & Res-50  & 11.0 & 49.3  \\
NMS-Loss \cite{luo2021nms} & - & 10.8 & -   \\
ACSP \cite{wang2020adapted} & Res-101 & 9.3 & 46.3 \\
PedesFormer \cite{hasan2022pedestrian} & - & \textbf{9.2} & \textbf{36.9}  \\
SOLIDER~\cite{chen2023beyond} & Swin-B & 9.7 & 39.4  \\ 
\hline
\specialrule{0em}{1pt}{1pt}
CLASP (CSP) & Swin-T & 9.9 & 40.3  \\
CLASP (CSP) & Swin-S & 9.5 & 38.6  \\
CLASP (CSP) & Swin-B & \textbf{9.2} & \underline{37.4}   \\
\specialrule{0em}{1pt}{1pt} 
\hline
\end{tabular}
\end{threeparttable}
\end{minipage}

\begin{minipage}{\columnwidth}
\centering
\caption{Comparison of GCR (Gradient Conflict Ratio),  EAD (Expert-Activation Divergence), and HM (Harmonic Mean).} \vspace{-0.15cm}
\label{tab:intertask_conflict}
\begin{threeparttable}
\begin{tabular}{@{}lccc@{}}
\toprule
Method & GCR $\downarrow$ & EAD $\uparrow$ & HM(ReID,Parsing) $\uparrow$ \\
\midrule
DINO \cite{caron2021emerging} (baseline) & 0.42  & -- & 67.75 \\
SOLIDER \cite{chen2023beyond} & 0.35 & -- & 73.59 \\
CLASP w/ standard MoE & 0.31 & 0.15 & 74.16 \\
\textbf{CLASP (ours)} & \textbf{0.27} & \textbf{0.34} & \textbf{76.72} \\
\bottomrule
\end{tabular}
\end{threeparttable}
\end{minipage}
\end{table}

\begin{table*}
  \centering
  \scriptsize
 \caption{Params, peak memory, training time per epoch, FLOPs and latency. Values are measured  under the same backbones and schedules with 8 A40 GPUs.} \label{compute_memory}
  \begin{tabular}{@{}lccccc@{}}
  \toprule
  Method & Params (M) & Peak Mem (GB) & Train time / epoch & FLOPs (G) & Latency (ms/img, FP32) \\
  \midrule
  DINO~\cite{caron2021emerging}      & 88 & 12.0 & 2.7h & 15.2 & 14.4 \\
  SOLIDER~\cite{chen2023beyond}   & 89 & 12.8 & 3.1h & 15.8 & 14.7 \\
  \textbf{CLASP} & 91 & 14.0 & 3.25h & 16.2 & 15.1 \\
  \bottomrule
  \end{tabular}
  \end{table*}

\begin{table*}
  \centering
  \scriptsize
  \caption{The number of epochs and time required for model (Swin-Transformer-base) convergence when fine-tuning Reid, human parsing, and attribute recognition downstream tasks on 4 v100 GPUs.} \label{downstream_convergence}
  \resizebox{\textwidth}{!}{
  \begin{tabular}{@{}lccccccccccc@{}}
  \toprule
  \multirow{2}{*}{Method} &\multicolumn{4}{c}{\textbf{ReID (Market1501)}}  & \multicolumn{3}{c}{\textbf{Human Parsing (LIP)}}  & \multicolumn{3}{c}{\textbf{Attribute Recognition (PA100k)}}  \\
  \cmidrule(lr){2-5} \cmidrule(lr){6-8} \cmidrule(lr){9-11}   & Training time & Training epochs & mAP $\uparrow$ & Rank1 $\uparrow$ & Training time & Training epochs & mIoU $\uparrow$ & Training time & Training epochs & mA $\uparrow$ \\
  \midrule
  DINO~\cite{caron2021emerging}      & 2.4h & 109 & 92.5 & 95.8 & 7.2h & 54 & 59.27 & 4.6h & 35 & 84.63 \\
  SOLIDER~\cite{chen2023beyond}   & 2.4h  & 106 & 93.9 & 96.9 & 5.5h & 41 & 60.50 & 3.2h & 31 & 86.37 \\
  \textbf{CLASP} & 2.2h & 103 & 94.6 & 97.1 & 4.3h & 36 & 61.74 & 2.05h & 27 & 87.91 \\
  \bottomrule
  \end{tabular}}
\end{table*}

\subsection{Comparison to State of the Arts}
 Tab. \ref{tab:reid} to \ref{tab:pedestrian_detection} demonstrate that the proposed CLASP method achieves competitive and even superior performance compared to state-of-the-art approaches across a wide range of human-centric tasks. \textbf{(1)} In person re-identification (ReID), CLASP (Swin-B) achieves the highest mAP of 94.6\% on Market1501 and 78.8\% on MSMT17, surpassing or closely matching state-of-the-art methods. 
 \textbf{(2)} For attribute recognition, CLASP attains the best mean accuracy (mA) across all evaluated datasets, highlighting its effectiveness in capturing human attributes under zero-shot settings. 
 \textbf{(3)} In human parsing, CLASP improves mIoU to 61.74\% on LIP, outperforming prior methods such as  SOLIDER. 
 \textbf{(4)} For person search, CLASP sets new state-of-the-art results, achieving 96.1\% mAP and 96.4\% Rank-1 on CUHK-SYSU, and 60.8\% mAP with 87.8\% Rank-1 on PRW, demonstrating strong capability in joint detection and identification. 
 \textbf{(5)} In pose estimation, it achieves 77.8\% AP and 82.2\% AR on the COCO dataset, indicating solid performance in keypoint-based tasks. 
 \textbf{(6)} For pedestrian detection, CLASP reduces MR$^{-2}$ to 9.2\% on CityPerson, outperforming SOLIDER (9.7\%), reflecting improved detection robustness.

As shown in Tab.~\ref{tab:intertask_conflict}, DINO \cite{caron2021emerging} and SOLIDER \cite{chen2023beyond} exhibit relatively high GCR, indicating considerable gradient conflicts among tasks. In contrast, our CLASP achieves the lowest GCR, which confirms its efficacy in mitigating inter-task gradient disagreements. With the integration of PC-MoE, different tasks activate distinct experts, as reflected by higher EAD. This expert's separation enables the effective decoupling of task-specific feature representations. Consequently, our method achieves superior overall performance, demonstrating a more
balanced and effective modeling of appearance and semantic features for person ReID and semantic parsing.

Tab.~\ref{compute_memory} provides a comparative analysis of computational costs. As shown, CLASP introduces only a marginal overhead—including +3M parameters, +6.3\% training time (equivalent to an increase of 0.55 hours per epoch), and +2.7\% inference latency—while achieving notable accuracy improvements over SOLIDER (Parsing +1.6\%, ReID mAP +2.3\%). This added cost stems mainly from the PC-MoE module, yet the overhead remains acceptable given the substantial performance gains across downstream tasks, demonstrating that CLASP achieves a favorable trade-off between efficiency and performance without relying on brute-force compute scaling. And Tab.~\ref{downstream_convergence} compares the convergence efficiency (in terms of epochs and time) of DINO, SOLIDER, and CLASP on downstream fine-tuning tasks. The results demonstrate that CLASP significantly reduces convergence time on downstream tasks,  while simultaneously achieving superior performance compared to both DINO and SOLIDER.
 
 These results collectively confirm that CLASP consistently delivers state-of-the-art or highly competitive performance across diverse tasks, highlighting its strong generalization ability and robustness in unified human-centric visual analysis.

\section{Conclusion}
In this paper, we propose CLASP, a novel self-supervised learning framework designed for human-centric visual analysis, effectively addressing the challenges of adapting to diverse downstream tasks. Unlike existing self-supervised approaches, CLASP leverages a general visual-language model to generate multi-level semantic labels, capturing both low-level body parts and high-level attributes. These pseudo part  and attribute semantic labels enable the model to  learn richer semantic representations. To further enhance adaptability, we propose a PC-MoE module, which integrates  channel-wise and global-level gating mechanisms  guided by task-specific learnable prompts. This design mitigates feature conflicts and improves generalization across tasks. Extensive experiments demonstrate  that CLASP achieves strong performance compared to state-of-the-art methods, particularly excelling in semantic human-centric tasks. 

\noindent\textbf{Broader impacts}. The proposed method significantly improves the effectiveness of human-centric tasks, making it highly applicable to security and surveillance scenarios. However, this enhanced performance also amplifies potential privacy risks, particularly in sensitive contexts. As such, it is imperative to establish robust regulatory frameworks governing the collection, usage, and management of human-related data to ensure ethical and responsible deployment.

\section*{Acknowledgment}

This work is partially supported by National Natural Science Foundation of China (NSFC): 62376259, 62576334 and 62306301.


\small \bibliographystyle{IEEEtran} \bibliography{main}

\begin{IEEEbiography}[{\includegraphics[width=1in,height=1.25in,clip,keepaspectratio]{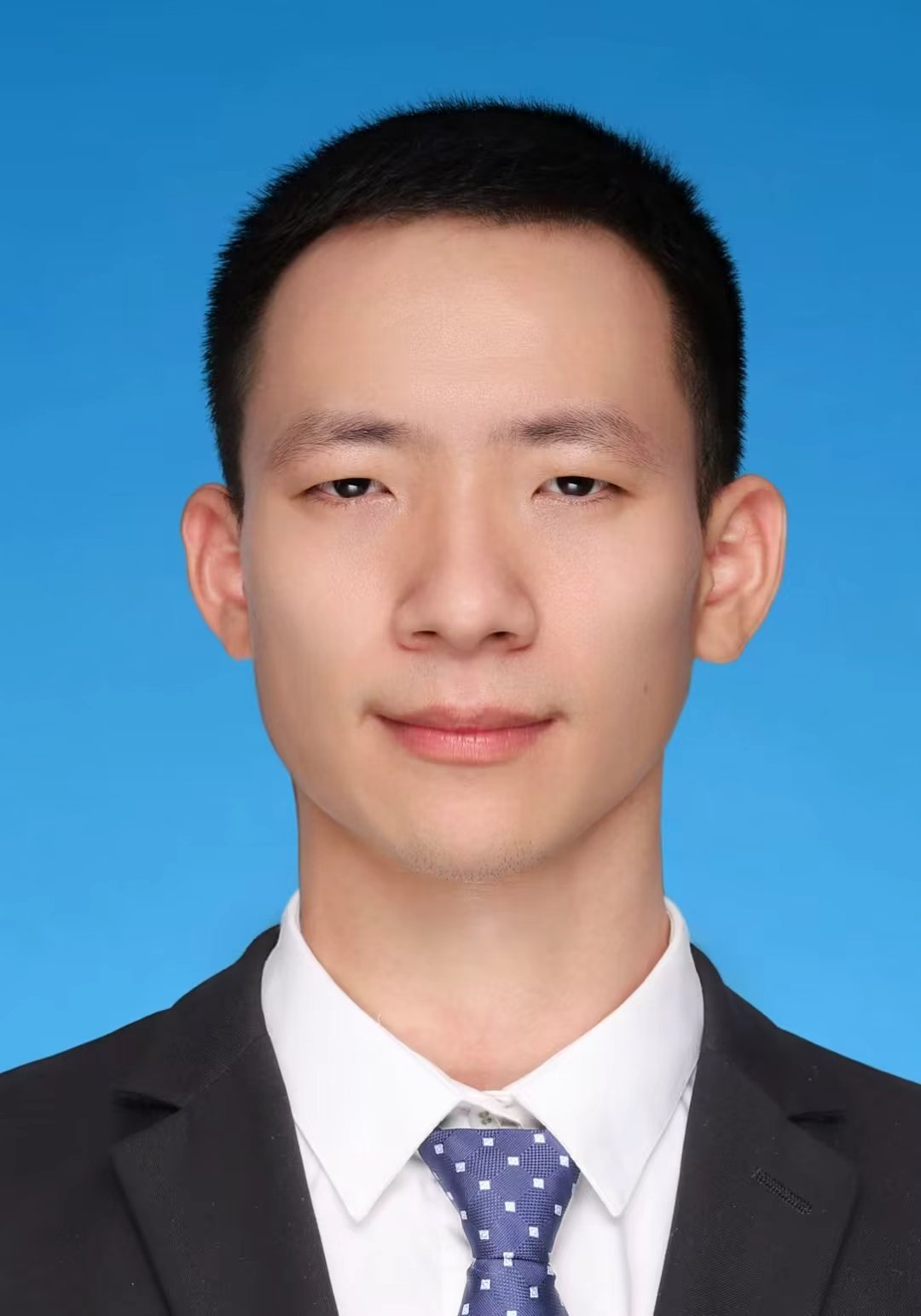}}]{Mingshuang Luo} 
received the BS degree in electric information science and technology from Xiangtan University in 2018; the MS degree in computer technology from the University of the Chinese Academy of Sciences (UCAS) in 2021. He has been pursuing the PhD degree at the Institute of Computing Technology (ICT), Chinese Academy of Sciences (CAS), since 2022. From 2021 to 2022, he worked at Xiaomi Group as a Speech Algorithm Researcher. His research interests are in computer vision, pattern recognition, and machine learning. He especially focuses on human motion generation, human animation generation and related research topics.
\end{IEEEbiography}

\begin{IEEEbiography}[{\includegraphics[width=1in,height=1.25in,clip,keepaspectratio]{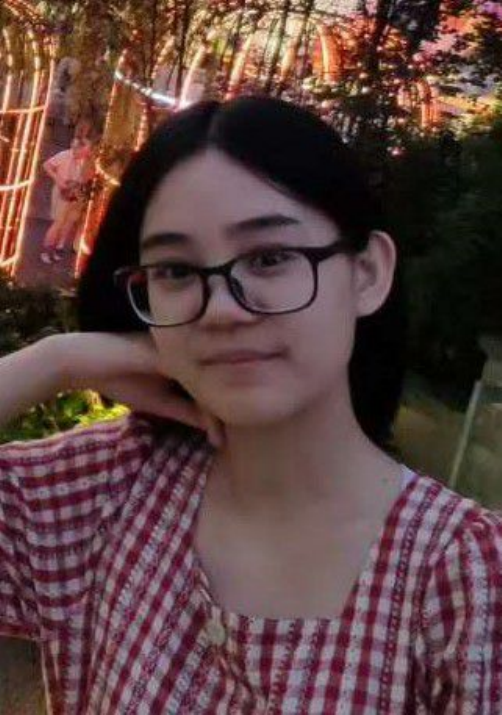}}]{Ruibing Hou} (Member, IEEE) 
received the BS degree in Northwestern Polytechnical University, Xi’an, China, in 2016. She received PhD degree in computer science from the Institute of Computing Technology, Chinese Academy of Sciences, Beijing, China, in 2022. She is currently a Associate Professor with the Institute of Computing Technology, Chinese Academy of Sciences. Her research interests are in machine learning and computer vision. She especially focuses on multimodal large language model and 3D humam modeling. 
\end{IEEEbiography}

\begin{IEEEbiography}[{\includegraphics[width=1in,height=1.25in,clip,keepaspectratio]{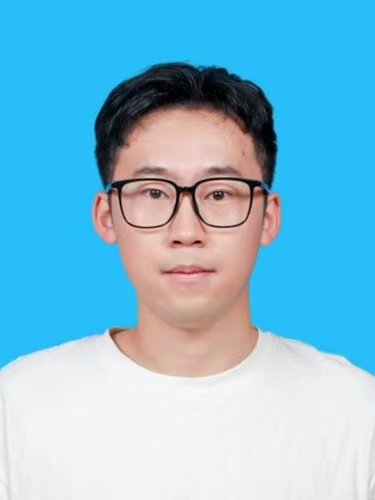}}]{Bo Chao} 
received the BS degree in computer science and technology from China Agricultural University in 2021, and the MS degree in electronic information from the Institute of Computing Technology (ICT), Chinese Academy of Sciences (CAS), in 2025. His research interests include computer vision, pattern recognition, and machine learning, with a particular focus on person re-identification and related research topics. 
\end{IEEEbiography}

\begin{IEEEbiography}[{\includegraphics[width=1in,height=1.25in,clip,keepaspectratio]{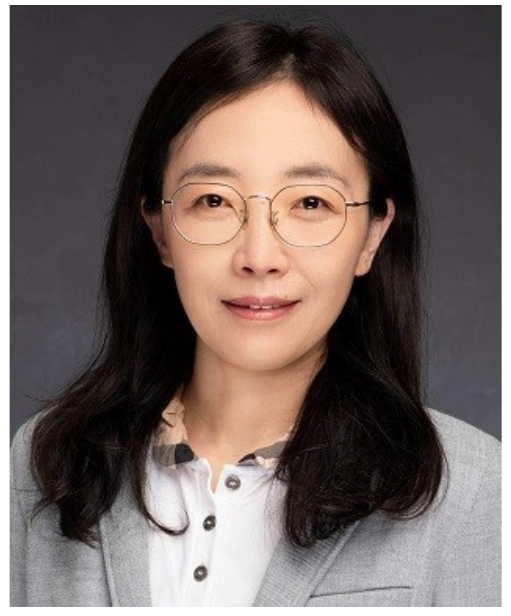}}]{Hong Chang} (Member, IEEE) 
received the bachelor’s degree in computer science from the Hebei University of Technology, Tianjin, China, in 1998, the MS degree in computer science from Tianjin University, Tianjin, in 2001, and the PhD degree in computer science from the Hong Kong University of Science and Technology, Kowloon, Hong Kong, in 2006. She was a research scientist with Xerox Research Centre Europe. She is currently a researcher with the Institute of Computing Technology, Chinese Academy of Sciences, Beijing, China. Her main research interests include algorithms and models in machine learning, and their applications in pattern recognition, computer vision and AI2Science.
\end{IEEEbiography}

\begin{IEEEbiography}[{\includegraphics[width=1in,height=1.25in,clip,keepaspectratio]{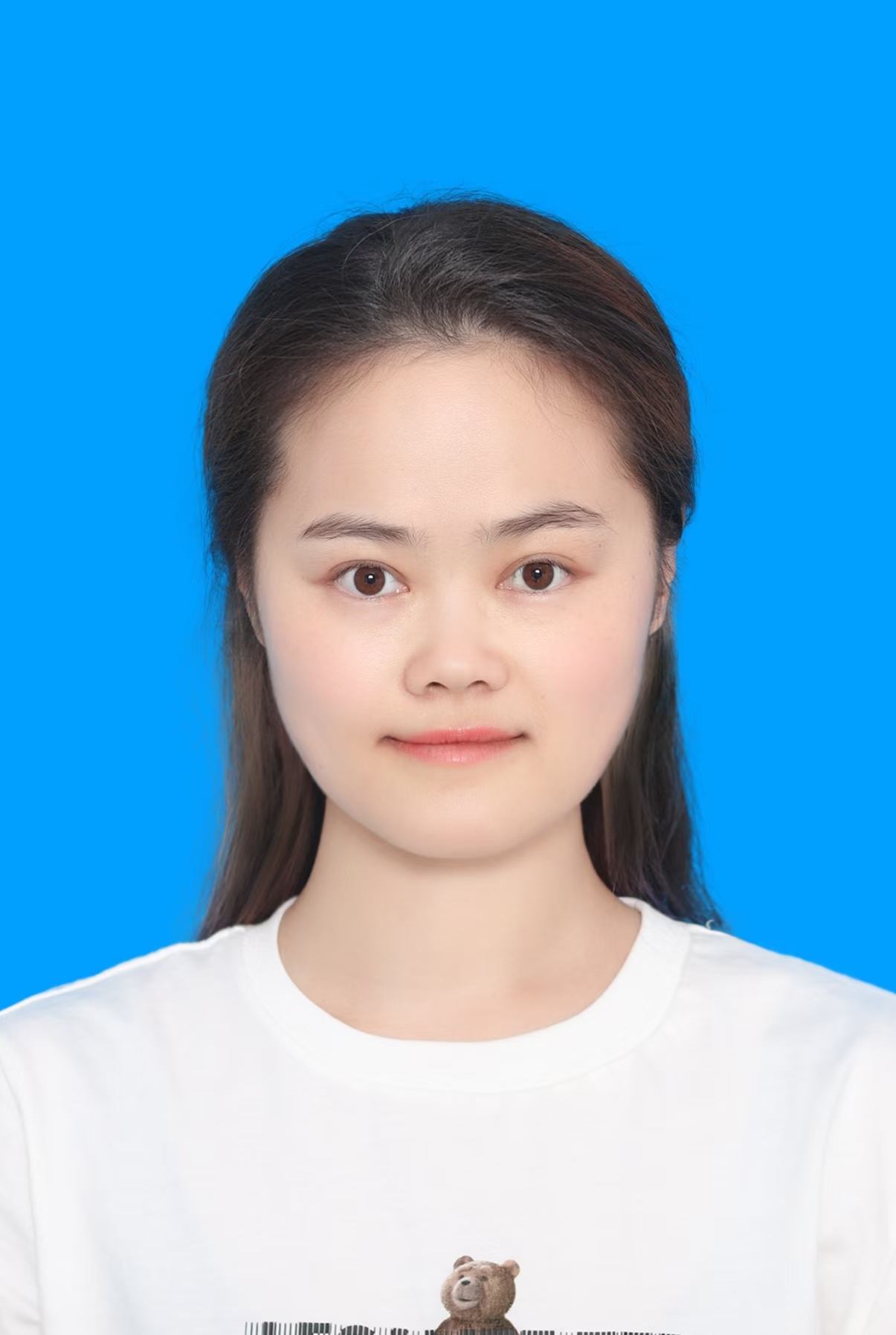}}]{Zimo Liu} received the Ph.D degrees in information and communication engineering in Dalian University of Technology, Dalian, China, in 2021. From 2017 to 2018, she conducted research as a joint PhD with the Vision Lab of Queen Mary University of London. She is currently an algorithm engineer in Peng Cheng Laboratory, Shenzhen. The current research interests include computer vision, model compression, low-light enhancement,and multi-modal learning.
\end{IEEEbiography}

\begin{IEEEbiography}[{\includegraphics[width=1in,height=1.25in,clip,keepaspectratio]{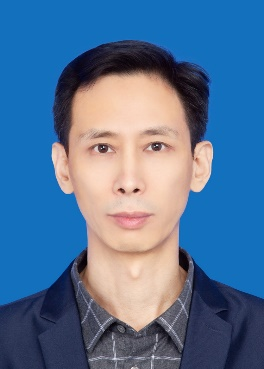}}]{Yaowei Wang} (Member, IEEE) is currently a professor with the Harbin Institute of Technology, Shenzhen and the Peng Cheng Laboratory. He is a recipient of the special government allowances of the State Council. He is the author of more than 100 high-impact papers in prestigious venues. His current research interests include multimedia content analysis and understanding, machine learning, and computer vision. He serves as the chair of the IEEE Digital Retina Systems Working Group and associate editor of the journal IEEE TCSVT, and also a member of IEEE, CIE, CCF, CSIG. He was the recipient of the second prize of the StateTechnological Invention Award in 2017, the first prize of the CIE Technology Invention Award in 2015, the first prize of the CIE Scientific and Technological Progress Award in 2022, and the special grand prize of Guangdong Technology Progress award in 2023.
\end{IEEEbiography}

\begin{IEEEbiography}[{\includegraphics[width=1in,height=1.25in,clip,keepaspectratio]{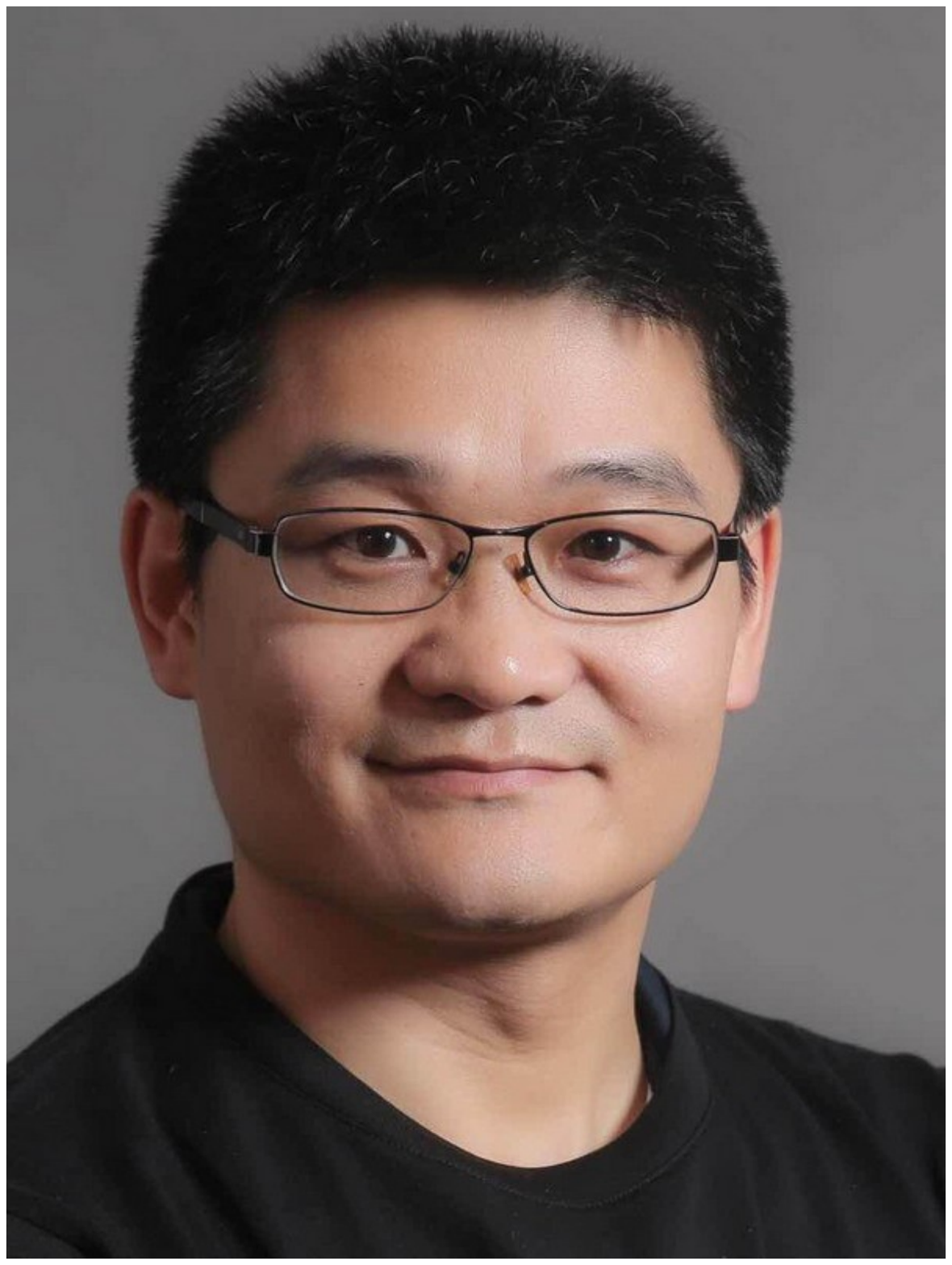}}]{Shiguang Shan} (Fellow, IEEE)
received the PhD degree in computer science from the Institute of Computing Technology, Chinese Academy of Sciences (CAS), Beijing, China, in 2004. Since 2010, he has been a full professor with the Institute of Computing Technology. He is currently the deputy director with the CAS Key Lab of Intelligent Information Processing. He has authored or coauthored more than 300 papers, with totally more than 20,000 Google scholar citations. His research interests include computer vision, pattern recognition, and machine learning. He was an area chairs for many international conferences including CVPR, ICCV, AAAI, IJCAI, ACCV, ICPR, and FG. He is/was associate editors of several journals including IEEE TRANSACTIONS ON IMAGE PROCESSING, Neurocomputing, Computer Vision and Image Understanding, and PRL. He was the recipient of the China’s State Natural Science Award in 2015 and China’s State S\&T Progress Award in 2005 for his research work.
\end{IEEEbiography}

\end{document}